\documentclass{article}

\PassOptionsToPackage{table}{xcolor}
\usepackage{opd_aha_preprint,times}

\usepackage{amsmath,amsfonts,bm}

\def\eqref#1{equation~\ref{#1}}

\def\1{\bm{1}}

\DeclareMathAlphabet{\mathsfit}{\encodingdefault}{\sfdefault}{m}{sl}
\SetMathAlphabet{\mathsfit}{bold}{\encodingdefault}{\sfdefault}{bx}{n}

\usepackage{amsmath}
\usepackage{amssymb}
\usepackage{dsfont}
\usepackage{booktabs}
\usepackage{array}
\usepackage{colortbl}
\usepackage{graphicx}
\usepackage{marvosym}
\usepackage{float}
\usepackage{flafter}
\usepackage{multirow}
\usepackage{placeins}
\usepackage{needspace}
\usepackage[normalem]{ulem}
\usepackage{etoolbox}
\usepackage[most]{tcolorbox}
\usepackage{tikz}
\usetikzlibrary{arrows.meta,positioning,fit}
\usepackage{hyperref}
\hypersetup{
  hidelinks,
  pdftitle={OPD-Aha: From Linguistic Momentum to Visual Reflection in Multimodal On-Policy Distillation},
  pdfauthor={Chenhao Qiu, Dawei Li, Yechao Zhang, Lei Gong, Zhen Tan},
  pdfsubject={},
  pdfkeywords={multimodal reasoning, on-policy distillation, visual reflection}
}
\usepackage{enumitem}

\definecolor{studentblue}{RGB}{218,235,249}
\definecolor{teacherorange}{RGB}{252,226,198}
\definecolor{gapgreen}{RGB}{220,240,222}
\definecolor{futurepurple}{RGB}{235,225,246}
\definecolor{reflectionpurple}{RGB}{117,103,174}
\definecolor{softgray}{RGB}{245,245,245}
\definecolor{tablehighlight}{RGB}{221,244,255}

\newcommand{\method}{OPD-Aha}

\newcommand{\stdobs}{I}
\newcommand{\privobs}{I^{+}}
\newcommand{\student}{p_{\theta}}

\newcommand{\privteacher}{p_{\phi}}

\AtBeginEnvironment{table}{\setlength{\belowcaptionskip}{6pt}}
\AtBeginEnvironment{table*}{\setlength{\belowcaptionskip}{6pt}}

\title{\hyphenpenalty=10000
OPD-Aha: From Linguistic Momentum to Visual Reflection in Multimodal On-Policy Distillation}

\author{%
\parbox[t]{\textwidth}{\centering\bfseries
\mbox{Chenhao Qiu}\quad
\mbox{Dawei Li$^{1}$}\quad
\mbox{Yechao Zhang}\quad
\mbox{Lei Gong$^{2}$}\quad
\mbox{Zhen Tan$^{3,\text{\href{mailto:ztan12@stevens.edu}{\Letter}}}$}}\\[0.6em]
\normalfont \parbox[t]{\textwidth}{\centering
\mbox{$^{1}$Arizona State University}\quad
\mbox{$^{2}$University of Virginia}\quad
\mbox{$^{3}$Stevens Institute of Technology}}}
\date{}

\begin{document}

\raggedbottom

\maketitle

\begin{abstract}
Privileged on-policy distillation improves multimodal reasoning by allowing a teacher to evaluate student trajectories using rich, training-only visual evidence. Both models score these trajectories while conditioning on the same student-generated prefix. When a student misinterprets an image early in a response, this accumulating erroneous rationale eventually pulls the teacher away from its visual evidence. The teacher and student converge on the same hallucination, causing standard cross-model supervision to collapse precisely where correction is most needed. We find that the teacher's visual corrective preference is not lost under this misleading agreement. Comparing the predictions of the identical teacher given the real image and a visual null reveals that the privileged evidence still pushes the model toward the correct interpretation. We introduce OPD-Aha, which reconstructs the distillation target directly from this isolated visual preference rather than relying on the fragile teacher-student discrepancy. This reconstructed target aggressively suppresses continuations that contradict the image. Trained with this objective, students learn to naturally interrupt their own flawed reasoning with reflection tokens such as \emph{wait} and \emph{actually}. After reflection, subsequent generation relies less on the accumulated erroneous text and more on the visual evidence. Correcting these trajectories mid-generation fundamentally alters the reasoning process, yielding broad and consistent improvements across diverse fine-grained perception and complex multimodal reasoning benchmarks.
Our code and models are available at \url{https://github.com/Echochef/OPD-Aha}.
\end{abstract}

\section{Introduction}
\label{sec:introduction}

On-policy distillation provides dense, token-level supervision on the exact
trajectories explored by a student
model~\citep{agarwal2024opd,gu2024minillm,zhao2026selfdistilled,jin2026eopd,li2026videoopd}. In multimodal reasoning~\citep{liu2023llava,li2023blip2,bai2023qwenvl},
privileged on-policy distillation strengthens this supervision by giving the
teacher access to richer, training-only visual evidence~\citep{vapnik2009privileged,lopezpaz2016unifying}, such as localized
high-resolution views, while the student continues to operate on its original
visual input~\citep{yuan2026visionopd,tian2026vicur,wei2026zoombench}. This privileged evidence is
used to evaluate the states visited by the student during its own rollout.
Accordingly, at every decoding step, the teacher combines its richer visual input
with the same student-generated linguistic prefix that defines the current student
state. As the rollout progresses, privileged visual supervision is therefore
delivered under an increasingly long context written by the student itself.

This coupling becomes problematic when the student makes an early perceptual
error~\citep{li2026cognitive}. Once this error enters the prefix~\citep{jiang2026trd,xu2026relay}, subsequent generation elaborates on an
interpretation that conflicts with the image~\citep{li2023pope,favero2024m3id,he2025vhr,guo2025lisa,chen2026resdec}. The
teacher must then evaluate its privileged visual evidence in the presence of an
increasingly strong linguistic context supporting the student's mistaken
interpretation. We show that as this erroneous text grows, its linguistic momentum
progressively dominates the teacher's predictions and marginalizes the visual
evidence. The teacher eventually abandons the visually grounded correction and
favors the student's hallucinated continuation. Standard privileged distillation
therefore loses its corrective signal precisely at the states where the student
most needs intervention.

Despite this apparent supervision collapse, we find that a robust visual
corrective preference survives in the teacher's predictions. We introduce OPD-Aha
to reconstruct the distillation target directly from this surviving signal. The
method exposes the hidden visual preference by evaluating the identical teacher
under the same shared prefix, varying only the visual input between the privileged
image and a visual null~\citep{leng2024vcd,favero2024m3id}. This intra-teacher
contrast strips away the linguistic momentum and isolates the pure effect of the
visual evidence, revealing that the privileged evidence continues to strongly
suppress the hallucinated continuation even when the teacher's overall token
distribution aligns with the student's erroneous reasoning. OPD-Aha translates
this real-null prediction difference into a regularized target distribution that
selectively suppresses image-inconsistent continuations while preserving a valid
language distribution from the privileged teacher. Distilling this reconstructed
target along the unchanged student rollout equips the student with a mechanism to
interrupt and re-anchor its generation on visual evidence.

Students trained with OPD-Aha learn to naturally interrupt their own flawed
reasoning, producing reflection tokens such as \emph{wait} and
\emph{actually}~\citep{deepseekai2025r1,zhou2025visualthinker} when their current
explanation conflicts with the image. We find that this behavior emerges not
because the reconstructed target directly increases the absolute probability of
reflection tokens, but because it suppresses the erroneous continuation more
strongly. This asymmetric suppression grants self-interruption a crucial relative
advantage precisely where correction is needed. We further examine how this
self-interruption alters the generation dynamics. After reflection, the student
relies less on its earlier erroneous explanation and draws more strongly on the
visual evidence when generating subsequent tokens. This restored visual reliance
yields consistent accuracy improvements across six fine-grained perception and
complex multimodal reasoning benchmarks.

Our contributions are as follows:
\begin{enumerate}[leftmargin=*]
    \item We identify a failure mode of privileged on-policy distillation:
    erroneous student prefixes can overwhelm privileged visual evidence,
    collapsing teacher--student supervision precisely when correction is most
    needed.
    \item We introduce OPD-Aha, which reconstructs supervision from an
    intra-teacher real--null contrast under the same shared student prefix. This
    contrast isolates a corrective visual preference that survives the collapse
    and enables self-interruption with renewed visual reliance.
    \item Training with OPD-Aha yields consistent accuracy improvements across
    fine-grained perception benchmarks and positive transfer to unseen multimodal
    reasoning tasks. More broadly, our results show that robust multimodal
    distillation requires preserving visual correction against the linguistic
    momentum of the student trajectory.
\end{enumerate}

\section{Student Prefixes Mask Privileged Supervision}
\label{sec:problem}

\subsection{Privileged Multimodal On-Policy Distillation}
\label{sec:privileged_multimodal_opd}

Privileged multimodal on-policy distillation exploits an asymmetry in visual
evidence between the teacher and student. Given a standard visual observation
$\stdobs$ and a textual query $x$, the trainable student $\student$ generates
a reasoning trajectory
$y \sim \student(\cdot \mid \stdobs, x)$.
A frozen teacher $\privteacher$ evaluates the student-generated states while
receiving additional training-only visual evidence $\privobs$, such as a
localized high-resolution view of the task-relevant region, that is unavailable
to the student.

Despite this visual asymmetry, the teacher and student are conditioned on the
same student-generated linguistic prefix
$h_t = (x, y_{<t})$ at every decoding step.
Standard privileged multimodal OPD~\citep{yuan2026visionopd} minimizes the token-level
divergence between their predictions over these shared contexts:
\begin{equation}
    \mathcal L_{\mathrm{MOPD}}(\theta)
    =
    \mathbb E_{y\sim\student(\cdot\mid\stdobs,x)}
    \left[
        \frac{1}{T}
        \sum_{t=1}^{T}
        D_{\mathrm{dist}}\!\left(
        \privteacher(\cdot\mid\privobs,h_t),
        \student(\cdot\mid\stdobs,h_t)
        \right)
    \right].
\label{eq:standard_mopd}
\end{equation}

Privileged visual evidence gives the teacher access to stronger visual grounding than the student. Standard OPD transfers this advantage through the teacher--student prediction discrepancy under their shared linguistic prefix.
We next examine how this shared prefix affects privileged supervision when the student's trajectory already contains an erroneous visual interpretation.

\subsection{Erroneous Student Prefixes Mask Privileged Supervision}
\label{sec:fixed_rollout_analysis}
We find that an erroneous student prefix can progressively override the
teacher's privileged visual evidence.
Although the teacher receives stronger visual evidence, its autoregressive
predictions remain conditioned on the same student-generated prefix, which may
already encode an interpretation that contradicts the image. To quantify this
effect, we retain progressively longer portions of failed Vision-OPD
trajectories~\citep{yuan2026visionopd} and measure the teacher's preference
between the correct answer and the student's realized incorrect answer under
each retained prefix. Their difference in length-normalized log-likelihood
defines a decision margin that indicates whether the teacher still favors a
visually grounded correction or the erroneous prefix has become dominant.

\begin{figure}[!htbp]
    \centering
    \includegraphics[width=0.9\linewidth]{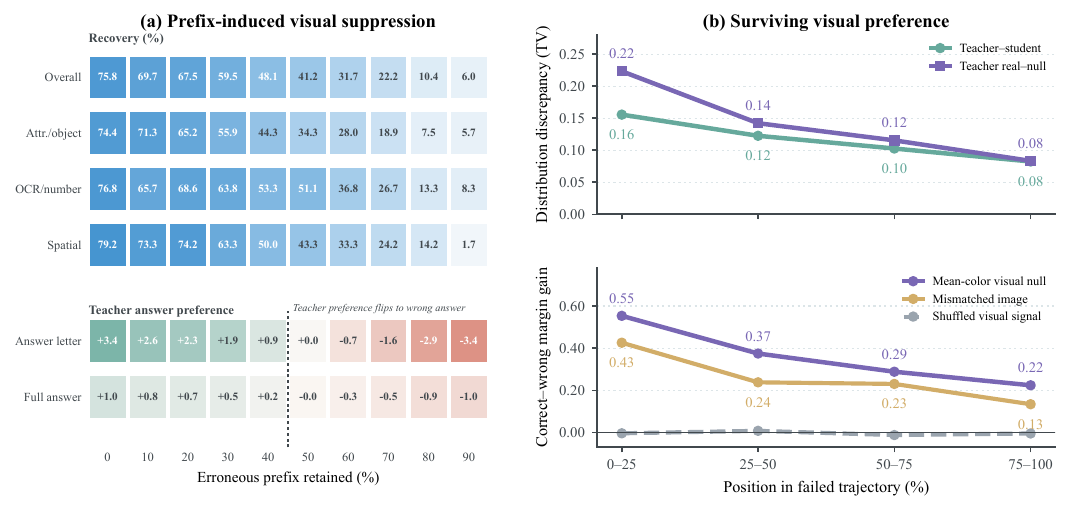}
    \caption{Erroneous prefixes mask privileged visual supervision. \textbf{(a)} As failed reasoning accumulates, the teacher's preference flips from the correct to the wrong answer. \textbf{(b)} The teacher--student discrepancy collapses, yet an intra-teacher visual contrast retains a strong preference for the correct answer. Token shuffling destroys this signal, confirming a token-specific visual preference.}
    \label{fig:hidden_visual_supervision_diagnostics}
\end{figure}

The privileged teacher reliably recovers the correct answer from short prefixes, but this ability drops sharply as erroneous reasoning accumulates (Figure~\ref{fig:hidden_visual_supervision_diagnostics}a). Near the midpoint of the response, its decision margin changes from positive to negative. The teacher no longer favors a visually grounded correction and instead prefers the student's realized wrong answer. As both models follow the same erroneous branch, their predictions converge and the cross-model discrepancy that drives standard privileged OPD disappears precisely where correction is most needed.

This convergence creates the impression that visual evidence has ceased to matter. We test this possibility by comparing two changes under the same student prefix: the difference between the teacher and student predictions, and the difference between the same teacher evaluated with privileged evidence and a visual null. The cross-model discrepancy collapses, while the real--null change remains substantial and continues to favor the correct answer (Figure~\ref{fig:hidden_visual_supervision_diagnostics}b). Replacing the visual null with a mismatched natural image preserves this direction, whereas shuffling the visual change across tokens destroys it. The surviving response is therefore a token-specific visual preference toward correction rather than undirected sensitivity to the input.

\FloatBarrier
\Needspace{12\baselineskip}

\section{Reconstructing Visual Supervision}
\label{sec:method}

\begin{figure}[!htbp]
    \centering
    \includegraphics[width=0.9\linewidth]{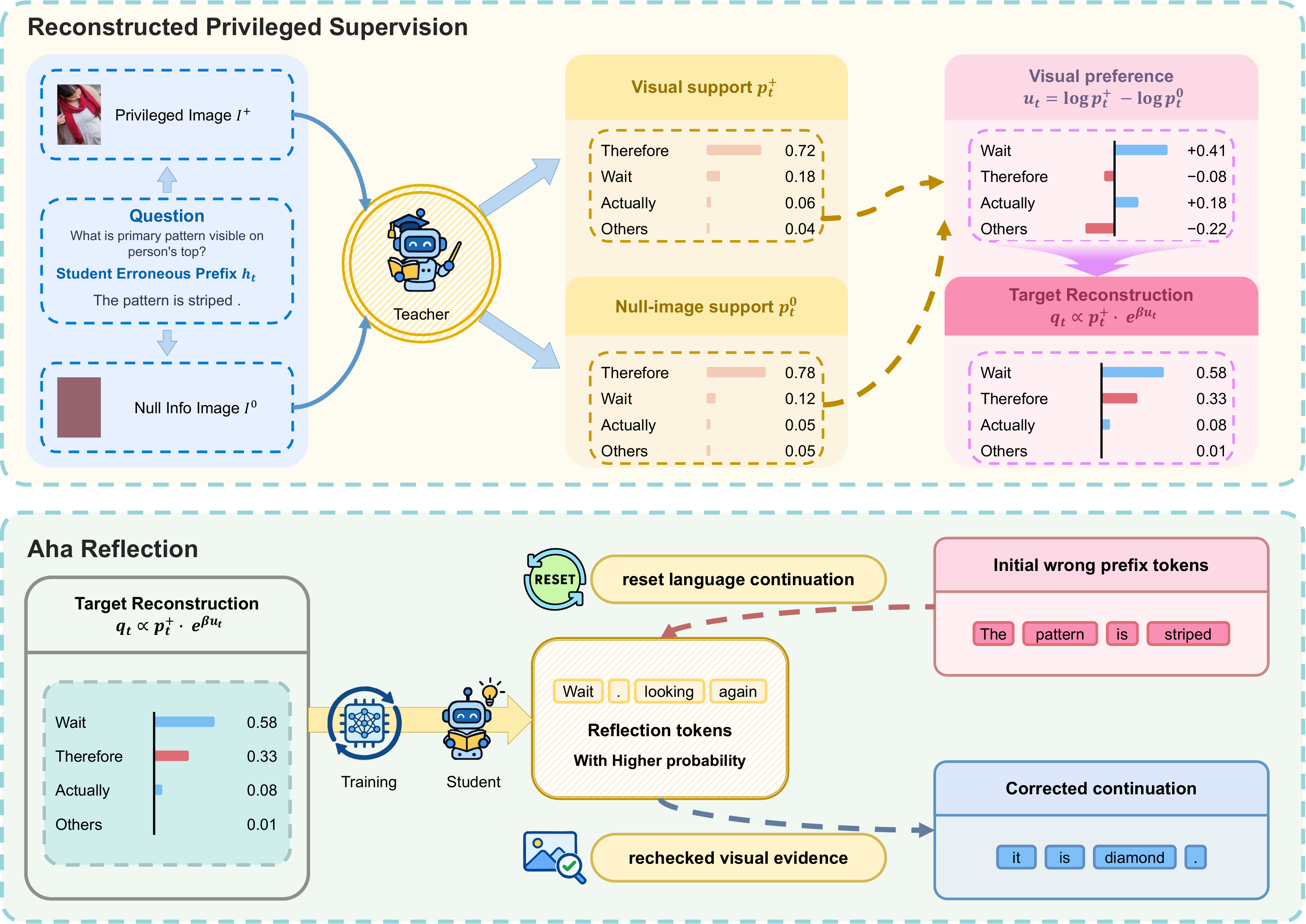}
    \caption{Overview of the \method{} framework.}
    \label{fig:method_overview}
\end{figure}

We introduce \method{} to reconstruct the visually grounded supervision that standard privileged OPD loses under an erroneous student prefix. As summarized in Figure~\ref{fig:method_overview}, the method abandons the unreliable cross-model comparison and instead extracts the surviving visual preference by contrasting the privileged teacher's predictions under real evidence and a visual null (Section~\ref{sec:matched_visual_counterfactual}). \method{} translates this isolated preference into a KL-regularized distillation target that selectively suppresses image-inconsistent continuations while preserving a valid language distribution (Section~\ref{sec:counterfactual_target}). Training the student on this reconstructed target transfers the visual correction along the unchanged rollout.

\subsection{Isolating Visual Preference}

\label{sec:matched_visual_counterfactual}

Teacher--student agreement under an erroneous prefix conceals how privileged evidence still shapes the teacher's token preferences. Isolating this surviving visual signal requires a comparison that does not depend on the student's distribution. We therefore evaluate the identical teacher under the same shared prefix and vary only the visual input~\citep{yang2024pensieve,zhao2025cicd}. We pair the privileged image $I^{+}$ with a visual null
\begin{equation}
    I^{0}=\mathcal N(I^{+}),
\end{equation}
where the transformation $\mathcal N$ removes the visual content while preserving the input dimensions. We instantiate $\mathcal N$ by replacing the image with its mean RGB color. Under the same student prefix $h_t$, we denote the teacher distributions for the privileged image and visual null, and the student distribution for the original image, respectively, by
\begin{equation}
    p_t^{+}(v)=p_{\phi}(v\mid I^{+},h_t),\quad
    p_t^{0}(v)=p_{\phi}(v\mid I^{0},h_t),\quad
    p_t^{S}(v)=p_{\theta}(v\mid I,h_t).
    \label{eq:null_teacher_distribution}
\end{equation}
The model and student prefix are shared across $p_t^{+}$ and $p_t^{0}$, so their difference isolates how privileged evidence changes the teacher's token preferences. We denote this visual preference signal by
\begin{equation}
    u_t(v)
    =
    \log p_t^{+}(v)
    -
    \log p_t^{0}(v).
    \label{eq:evidence_preference_change}
\end{equation}
A positive $u_t(v)$ indicates that the evidence raises the teacher's preference for token $v$, while a negative value indicates that the evidence suppresses it. Even when $p_t^{+}$ and $p_t^{S}$ are nearly indistinguishable, $u_t$ can remain nonzero and reveal the visual preference hidden by their agreement.

Because $p_t^{+}$ and $p_t^{0}$ use the same teacher and student prefix, $u_t$
attributes the prediction change to visual input rather than differences
between models. A nonzero $u_t$ is not necessarily corrective~\citep{yin2025mirage}. It becomes
useful when visual evidence favors tokens that leave the erroneous continuation
over tokens that sustain it. The signal provides a signed direction over
tokens, but it is not a normalized target and does not determine how far the
target should move from the privileged teacher.

\subsection{Target Reconstruction from Visual Preference}
\label{sec:counterfactual_target}

The visual preference signal $u_t(v)$ isolates a corrective direction over vocabulary tokens, but it is not a normalized target distribution. Following this direction alone would discard the structural knowledge of the language model, while directly distilling the privileged teacher $p_t^{+}$ would retain the distribution already dominated by the erroneous student prefix. We reconstruct a valid distillation target by balancing a candidate distribution $q\in\Delta(\mathcal V)$ between its alignment with the visual preference and its proximity to the original privileged teacher. This trade-off defines a KL-regularized objective:
\begin{equation}
    q_t
    =
    \underset{q\in\Delta(\mathcal V)}{\arg\max}
    \left\{
        \beta\,\mathbb E_{v\sim q}\!\left[u_t(v)\right]
        -
        D_{\mathrm{KL}}\!\left(q\,\|\,p_t^{+}\right)
    \right\}.
\label{eq:evidence_regularized_target}
\end{equation}

The coefficient $\beta\geq 0$ controls the strength of the visual correction. Setting $\beta=0$ leaves the base teacher distribution unchanged and recovers standard privileged OPD. Solving this objective applies an exponential tilt to the privileged target:
\begin{equation}
\begin{aligned}
    q_t(v)
    &=
    \frac{
        p_t^{+}(v)\exp\!\left(\beta u_t(v)\right)
    }{
        \sum_{w\in\mathcal V}
        p_t^{+}(w)\exp\!\left(\beta u_t(w)\right)
    }.
\end{aligned}
\label{eq:counterfactual_target}
\end{equation}

Substituting the definition of $u_t$ expands this solution into its component distributions:
\begin{equation}
    q_t(v)
    =
    \operatorname{softmax}
    \left(
        (1+\beta)\log p_t^{+}
        -
        \beta\log p_t^{0}
    \right)_v.
\label{eq:counterfactual_target_closed}
\end{equation}

This expanded form reveals the mechanics of target reconstruction. The base probability $p_t^{+}$ preserves the teacher's complete belief under real evidence, ensuring that the target remains a valid language distribution. The likelihood ratio $(p_t^{+}/p_t^{0})^{\beta}$ then selectively amplifies or suppresses tokens based on how the visual evidence changes the teacher's preference. The target moves away from standard privileged OPD only along the directions favored by the visual input.

The effect of this selective amplification becomes clear when evaluating the relative odds of two competing tokens $v$ and $w$:
\begin{equation}
    \log\frac{q_t(v)}{q_t(w)}
    =
    \log\frac{p_t^{+}(v)}{p_t^{+}(w)}
    +
    \beta\bigl(u_t(v)-u_t(w)\bigr).
\label{eq:target_pairwise_odds}
\end{equation}

This competition governs whether the student continues its current explanation or interrupts itself. If $w$ is an image-inconsistent continuation heavily favored by the inherited prefix, the base teacher preference $\log(p_t^{+}(v)/p_t^{+}(w))$ will strongly support $w$. Reconstruction can overturn this preference and promote a reflection token $v$ if the visual evidence suppresses the continuation strongly enough to make the second term dominant. The parameter $\beta$ determines how much visual separation is required to cross this threshold (Appendix~\ref{sec:target_optimality_appendix}).

The reconstructed target $q_t$ replaces the standard privileged teacher in the distillation divergence:
\begin{equation}
    \ell_t
    =
    D_{\mathrm{dist}}\!\left(q_t,p_t^{S}\right).
\label{eq:counterfactual_distillation_loss}
\end{equation}

The training objective averages this loss over all valid response positions $M_t$:
\begin{equation}
    \mathcal L_{\method}
    =
    \frac{
        \sum_t M_t\,\ell_t
    }{
        \sum_t M_t
    },
    \label{eq:method_objective}
\end{equation}

The student optimizes this objective along its own unchanged rollout. The visual preference is transferred entirely through the adjusted target probabilities, allowing the student to learn visually grounded corrections without requiring auxiliary visual inputs during inference.

\section{Emergent Reflection from Reconstructed Supervision}
\label{sec:emergent_reflection}

\subsection{Reconstruction Sustains Corrective Supervision}
\label{sec:mechanistic_validation}

We test whether target reconstruction preserves support for the correct answer as erroneous reasoning accumulates. Starting from failed student responses, we retain progressively longer prefixes and compare the correct-answer probability under the standard privileged target $p_t^{+}$ and the reconstructed target $q_t$ at the same prefix state. We also vary reconstruction strength $\beta$ during training and track response length and final benchmark accuracy to examine how changes in correct-answer probability relate to the student's generation and performance.

\begin{figure}[!htbp]
    \centering
    \includegraphics[width=0.99\textwidth]{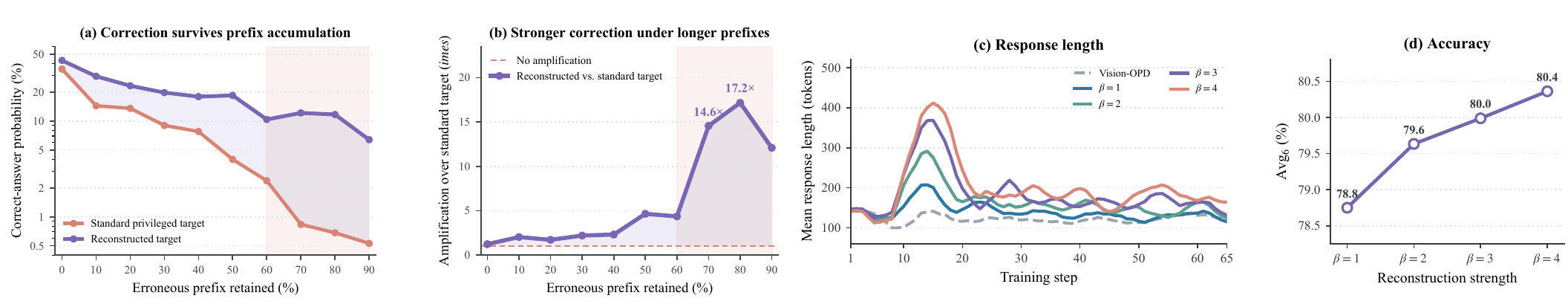}
    \caption{Late-prefix target recovery and training trends. \textbf{(a, b)} As hallucinated text accumulates, standard privileged distillation abandons the correct answer, whereas target reconstruction isolates the surviving visual preference to selectively amplify the correct token. \textbf{(c, d)} Stronger reconstruction induces a larger transient increase in response length before stabilizing, ultimately yielding consistent improvements in final accuracy.}
    \label{fig:revision_mechanism_sequence}
\end{figure}

Under the same short prefix, we find that both targets assign similar probabilities to the correct answer. As the erroneous prefix accumulates, standard privileged supervision sharply abandons the correct answer. The reconstructed target instead isolates the surviving visual preference, maintaining over an order of magnitude higher correct-answer probability at late-prefix states (Figure~\ref{fig:revision_mechanism_sequence}a,b). Support for the correct answer therefore remains stronger under the reconstructed target even after the student's trajectory has deviated.

This sustained visual correction alters the student's generation dynamics during training. Students trained with stronger reconstruction exhibit a larger transient increase in response length before their trajectories stabilize (Figure~\ref{fig:revision_mechanism_sequence}c). Across the same range of reconstruction strengths, overall accuracy averaged across the evaluated visual benchmarks improves consistently (Figure~\ref{fig:revision_mechanism_sequence}d). Preserving support for the correct answer along erroneous prefixes is thus accompanied by changes in generation and improved final performance.

\Needspace{8\baselineskip}

\subsection{Suppressing Erroneous Continuations Elicits Reflection}
\label{sec:reflection_emergence}

\begin{figure}[!htbp]
    \centering
    \begin{minipage}[t]{0.55\textwidth}
        \centering
        \vspace{0pt}
        \begin{tikzpicture}
            \node[inner sep=0] (tokenpanel) {\includegraphics[width=\linewidth]{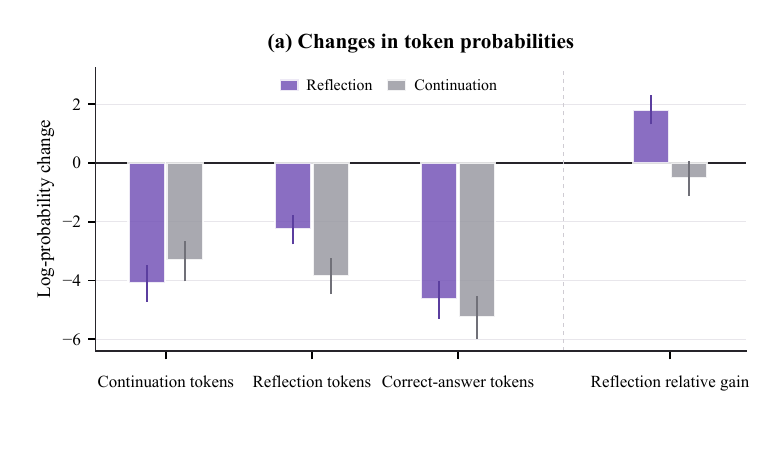}};
            \fill[white] ([xshift=22pt,yshift=-18pt]tokenpanel.north west) rectangle ([xshift=-2pt,yshift=-1pt]tokenpanel.north east);
            \node[anchor=base,inner sep=0pt,font=\fontsize{7}{8}\selectfont\bfseries] at ([xshift=10pt,yshift=-8pt]tokenpanel.north) {(a) Changes in token probabilities};
        \end{tikzpicture}
    \end{minipage}\hspace{0.02\textwidth}
    \begin{minipage}[t]{0.40\textwidth}
        \centering
        \vspace{0pt}
        \begin{tikzpicture}
            \node[inner sep=0] (dynamicspanel) {\includegraphics[width=\linewidth]{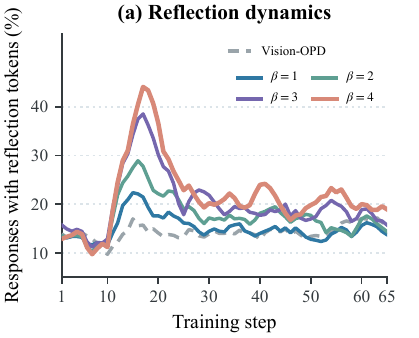}};
            \fill[white] ([xshift=22pt,yshift=-18pt]dynamicspanel.north west) rectangle ([xshift=-2pt,yshift=-1pt]dynamicspanel.north east);
            \node[anchor=base,inner sep=0pt,font=\fontsize{7}{8}\selectfont\bfseries] at ([xshift=6pt,yshift=-8pt]dynamicspanel.north) {(b) Reflection during training};
        \end{tikzpicture}
    \end{minipage}
    \caption{Token probabilities and reflection during training. \textbf{(a)} Target reconstruction selectively suppresses erroneous continuations before reflection, granting reflection tokens a relative advantage absent at non-reflection positions. \textbf{(b)} Stronger reconstruction increases the frequency of reflection tokens during training before stabilizing.}
    \label{fig:reflection_emergence_diagnostic}
\end{figure}
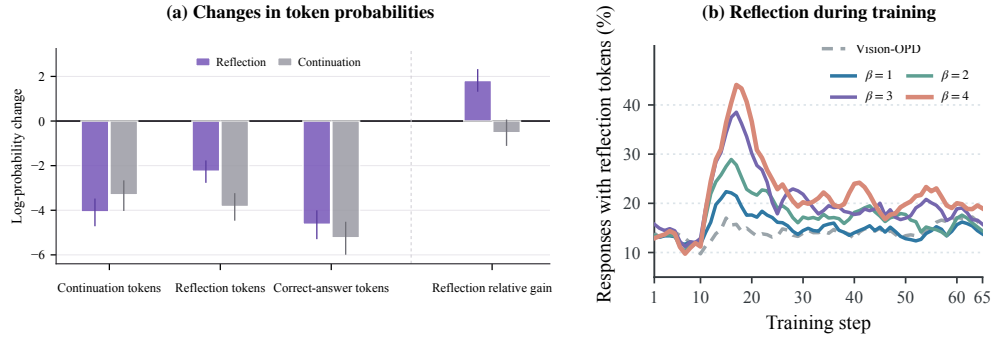

We examine whether reconstruction promotes reflection by raising its probability or suppressing the erroneous continuation more strongly. We compare how reconstruction changes token probabilities at positions immediately before reflection and at matched positions in responses without reflection (Figure~\ref{fig:reflection_emergence_diagnostic}a). For each token category, we measure the log-probability change from $p_t^{+}$ to $q_t$, using its total probability. We evaluate the relative log-probability gain of reflection tokens over continuation tokens to quantify the competitive advantage of self-interruption. To examine how these local probability adjustments translate into generation behavior, we track the overall frequency of reflection tokens throughout training across reconstruction strengths.

At these pre-reflection states, we find that reflection gains a relative advantage because reconstruction suppresses the erroneous continuation more strongly, even as the total probability of reflection tokens decreases. The correct-answer probability also decreases at these positions (Figure~\ref{fig:reflection_emergence_diagnostic}a). The visual preference thus discourages continuing the image-inconsistent explanation without directly rewarding reflection. Matched positions without reflection do not show the same increase in reflection tokens' relative probability.

This relative advantage is accompanied by more frequent reflection during training. As reconstruction strength increases, reflection tokens become more frequent during the transient phase and remain elevated for the stronger settings after stabilization, while Vision-OPD shows no comparable transition (Figure~\ref{fig:reflection_emergence_diagnostic}b). The student thus learns to interrupt its explanation when visual evidence reduces support for continuing it.

\Needspace{8\baselineskip}

\subsection{Reflection Restores Visual Reliance}
\label{sec:rollout_regrounding}

We examine whether reflection is followed by a shift from textual to visual reliance by aligning generated responses at their first reflection token and comparing subsequent tokens with matched positions in responses without reflection. At each position, we measure predictive support from the accumulated student prefix and support gained from visual evidence, tracking how their balance evolves after self-interruption.

\begin{figure}[H]
    \centering
    \includegraphics[width=0.85\textwidth]{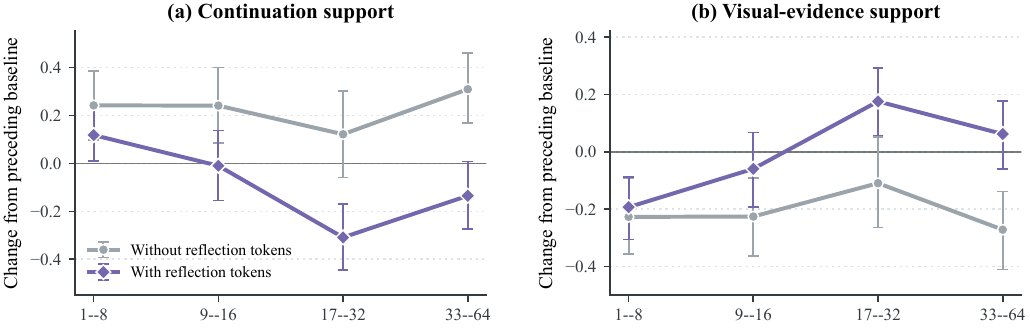}
    \caption{Visual reliance after reflection. \textbf{(a)} Support from the accumulated student prefix decreases after reflection. \textbf{(b)} Support gained from visual evidence rises after a short delay.}
    \label{fig:revision_cue_dynamics}
\end{figure}

We observe that this balance shifts toward visual evidence after reflection compared with matched positions in responses without reflection. Support from the accumulated student prefix decreases, followed by an increase in visual support after a short delay (Figure~\ref{fig:revision_cue_dynamics}). Reflection therefore marks a transition toward visual evidence, with a temporal gap between self-interruption and the recovery of visual support.

\FloatBarrier
\Needspace{8\baselineskip}

\section{Experiments}
\label{sec:experiments}

\subsection{Performance on Fine-Grained Visual Reasoning}
\label{sec:main_results}

Reconstructing the distillation target from the teacher's visual preference consistently translates into stronger downstream visual reasoning. \method{} outperforms Vision-OPD across all six fine-grained, high-resolution, and real-world benchmarks at both evaluated model scales (Table~\ref{tab:main_results_structure}). Rather than relying on fragile cross-model discrepancies, the reconstructed supervision provides a broad advantage, yielding a $4.0\%$ absolute increase in accuracy at 4B and a $3.2\%$ increase at 9B.

\begin{table}[H]
\caption{\method{} consistently outperforms standard privileged distillation and zero-shot baselines across diverse visual reasoning benchmarks. All \method{} results use $\beta=4$.}
\label{tab:main_results_structure}
\centering
\begingroup
\scriptsize
\renewcommand{\arraystretch}{0.92}
\setlength{\tabcolsep}{3.5pt}
\resizebox{\textwidth}{!}{%
\begin{tabular}{l|c|cc|cc|cc|c}
\toprule
\textbf{Models} & \textbf{Size}
& \multicolumn{2}{c|}{\textbf{Fine-Grained}}
& \multicolumn{2}{c|}{\textbf{High-Resolution}}
& \multicolumn{2}{c|}{\textbf{Real-World}}
& \textbf{Avg$_6$} \\
\cmidrule(lr){3-4}\cmidrule(lr){5-6}\cmidrule(lr){7-8}
& & \textbf{V$^\star$} & \textbf{Zoom}
& \textbf{HR-4K} & \textbf{HR-8K}
& \textbf{MME-EN} & \textbf{MME-CN} & \\
\midrule
\multicolumn{9}{l}{\textbf{\textit{General-Purpose Models}}} \\
Gemini 3.1 Pro (Google DeepMind, 2026) & --
& 88.2 & 61.6 & 88.8 & 85.8
& 75.1 & 72.5 & 78.7 \\
GPT-5.4 (OpenAI, 2026) & --
& 81.4 & 55.1 & 85.3 & 77.9
& 74.2 & 70.8 & 74.1 \\
Kimi-K2.6 (Moonshot AI, 2026) & 1T
& 86.3 & 53.8 & 82.7 & 78.7
& 69.7 & 66.4 & 72.9 \\
\midrule
\multicolumn{9}{l}{\textbf{\textit{``Thinking-with-Images'' Agentic Models}}} \\
DeepEyes (Zheng et al., 2026) & 7B
& 82.6 & 46.2 & 75.5 & 70.0
& 64.0 & 62.4 & 66.8 \\
Thyme-RL (Zhang et al., 2025a) & 7B
& 78.5 & 46.2 & 78.9 & 70.9
& 64.2 & 61.4 & 66.7 \\
DeepEyesV2 (Hong et al., 2026) & 7B
& 78.0 & 46.1 & 79.8 & 72.6
& 64.3 & 61.9 & 67.1 \\
SenseNova-MARS (Chng et al., 2026) & 8B
& 88.4 & 49.0 & 85.0 & 77.3
& 67.3 & 65.7 & 72.1 \\
\midrule
Qwen3.5 (Qwen Team, 2026) & 4B
& 82.7 & 49.1 & 86.5 & 81.5
& 59.8 & 61.1 & 70.1 \\
GRPO (Shao et al., 2024) & 4B
& 85.2 & 57.3 & 78.7 & 75.4
& 70.9 & 68.5 & 72.7 \\
V-Zero (Sun et al., 2026) & 4B
& 78.0 & 56.5 & 84.8 & 80.9
& 73.2 & 71.1 & 74.1 \\
Vision-OPD (Yuan et al., 2026) & 4B
& 89.0 & 59.5 & 83.4 & 80.5
& 74.6 & 71.6 & 76.4 \\
\rowcolor{tablehighlight}\textbf{\method{} (Ours)} & 4B
& \textbf{93.7} & \textbf{62.5} & \textbf{88.6} & \textbf{84.8}
& \textbf{77.7} & \textbf{74.9} & \textbf{80.4} \\
\midrule
Qwen3.5 (Qwen Team, 2026) & 9B
& 85.3 & 52.2 & 86.0 & 81.6
& 71.3 & 67.5 & 74.0 \\
GRPO (Shao et al., 2024) & 9B
& 87.2 & 57.5 & 86.0 & 83.0
& 73.3 & 69.2 & 76.0 \\
V-Zero (Sun et al., 2026) & 9B
& 89.2 & 59.4 & 85.8 & 83.3
& 75.7 & 70.6 & 77.3 \\
Vision-OPD (Yuan et al., 2026) & 9B
& 91.1 & 62.5 & 87.0 & 86.1
& 73.2 & 69.5 & 78.2 \\
\rowcolor{tablehighlight}\textbf{\method{} (Ours)} & 9B
& \textbf{94.8} & \textbf{63.9} & \textbf{88.9} & \textbf{87.5}
& \textbf{78.3} & \textbf{75.2} & \textbf{81.4} \\
\bottomrule
\end{tabular}
}
\endgroup
\end{table}

\Needspace{8\baselineskip}

\subsection{Comparison of Reconstruction Signals}
\label{sec:signal_source_ablation}

We compare targets reconstructed from the standard teacher--student discrepancy ($\log p_t^+ - \log p_t^S$) and the intra-teacher real--null difference ($\log p_t^+ - \log p_t^0$) to verify that the downstream gains originate from the isolated visual preference. Evaluated at the same reconstruction strength ($\beta=1$), amplifying the cross-model disagreement yields a marginal average increase over the standard privileged target, rising from 76.4\% to 76.9\%. Anchoring the reconstruction to the real--null difference produces consistent improvements across all six benchmarks and reaches an average accuracy of 78.8\% (Table~\ref{tab:signal_source_ablation}).

\begin{table}[H]
\caption{The same-teacher real--null difference provides a more effective reconstruction signal than teacher--student disagreement. All reconstructed targets use $\beta=1$.}
\label{tab:signal_source_ablation}
\centering
\setlength{\tabcolsep}{4.2pt}
\resizebox{\textwidth}{!}{%
\begin{tabular}{l|c|cc|cc|cc|c}
\toprule
\textbf{Target} & \textbf{Reference} & \textbf{V$^\star$} & \textbf{Zoom} & \textbf{HR-4K} & \textbf{HR-8K} & \textbf{MME-EN} & \textbf{MME-CN} & \textbf{Avg$_6$} \\
\midrule
Standard privileged target & --
& 89.0 & 59.5 & 83.4 & 80.5 & 74.6 & 71.6 & 76.4 \\
Teacher--student discrepancy & $p_t^{S}$
& 91.6 & 59.2 & 84.5 & 79.3 & 74.9 & 71.8 & 76.9 \\
\rowcolor{tablehighlight}\textbf{Teacher real--null difference} & $p_t^{0}$
& \textbf{93.2} & \textbf{61.5} & \textbf{86.0} & \textbf{82.0} & \textbf{76.5} & \textbf{73.2} & \textbf{78.8} \\
\bottomrule
\end{tabular}
}
\end{table}

\Needspace{8\baselineskip}

\subsection{Generalization to Multimodal Reasoning}
\label{sec:reasoning_generalization}

We evaluate zero-shot transfer on MathVerse~\citep{zhang2024mathverse}, MathVista~\citep{lu2024mathvista}, MathVision~\citep{wang2024mathvision}, WeMath~\citep{qiao2025wemath}, and DynaMath~\citep{zou2025dynamath}.
The benefits of target reconstruction extend robustly beyond the fine-grained perception domain used during training. When evaluated zero-shot on complex multimodal reasoning tasks, Vision-OPD degrades the 4B student's reasoning capabilities, pushing its performance below the unaligned Base model across all eight metrics. This broad regression indicates that its supervision is shaped by domain-specific linguistic habits from the training data, which fail to transfer out of distribution.

\begin{table}[H]
\caption{Target reconstruction enables zero-shot generalization to complex multimodal reasoning tasks. \method{} improves over the base model on all evaluated reasoning metrics. All \method{} results use $\beta=4$.}
\label{tab:reasoning_generalization}
\centering
\resizebox{\textwidth}{!}{%
\begin{tabular}{l|c|ccc|ccc|cc}
\toprule
\multirow{2}{*}{\textbf{Method}} & \multirow{2}{*}{\textbf{Size}} & \multirow{2}{*}{\textbf{MathVerse}} & \multirow{2}{*}{\textbf{MathVista}} & \multirow{2}{*}{\textbf{MathVision}} & \multicolumn{3}{c|}{\textbf{WeMath}} & \multicolumn{2}{c}{\textbf{DynaMath}} \\
\cmidrule(lr){6-8}\cmidrule(lr){9-10}
& & & & & \textbf{Row Acc.} & \textbf{Strict} & \textbf{Loose} & \textbf{Avg.} & \textbf{Worst} \\
\midrule
Qwen3.5 & 4B & 73.6 & 78.4 & 54.9 & 81.2 & 62.9 & 79.5 & 71.1 & 43.1 \\
GRPO & 4B & 70.0 & 74.2 & 46.1 & 74.5 & 52.5 & 69.6 & 64.6 & 31.5 \\
Vision-OPD & 4B & 71.0 & 74.4 & 50.3 & 80.0 & 59.8 & 77.9 & 67.3 & 38.5 \\
\rowcolor{tablehighlight}\textbf{\method{} (Ours)} & 4B & \textbf{75.7} & \textbf{80.0} & \textbf{55.6} & \textbf{84.7} & \textbf{69.4} & \textbf{81.6} & \textbf{72.6} & \textbf{44.5} \\
\midrule
Qwen3.5 & 9B & 78.3 & 80.9 & 45.5 & 83.7 & 68.7 & 81.1 & 71.2 & 46.8 \\
GRPO & 9B & 78.4 & 81.7 & 46.4 & 86.6 & 71.8 & \textbf{86.9} & 72.0 & 46.3 \\
Vision-OPD & 9B & 77.8 & \textbf{82.8} & 46.7 & 86.0 & 71.1 & 85.7 & 71.9 & 44.7 \\
\rowcolor{tablehighlight}\textbf{\method{} (Ours)} & 9B & \textbf{78.5} & 82.6 & \textbf{47.4} & \textbf{86.8} & \textbf{72.5} & 84.9 & \textbf{72.8} & \textbf{47.7} \\
\bottomrule
\end{tabular}
}
\end{table}

\method{} reverses this degradation and yields consistent positive transfer over the Base model across all unseen reasoning metrics at both scales (Table~\ref{tab:reasoning_generalization}). By distilling the isolated visual preference rather than the teacher's confounded output distribution, target reconstruction equips the student with a generalizable mechanism to re-anchor its generation on visual evidence. Consequently, the ability to suppress image-inconsistent continuations remains effective even when the underlying task semantics and reasoning complexity fundamentally change.

\FloatBarrier
\Needspace{8\baselineskip}

\section{Related Work}
\label{sec:related_work}

\paragraph{On-policy distillation with privileged information.}
Knowledge distillation transfers predictive structure from a stronger teacher to a deployable student, while learning with privileged information permits additional signals during training that are absent at inference~\citep{hinton2015distilling,vapnik2009privileged}.
Generalized distillation connects these paradigms by casting privileged information as teacher supervision~\citep{lopezpaz2016unifying}.
In imitation learning, DAgger addresses the distribution shift caused by a learner's own predictions by gathering supervision on the states it visits~\citep{ross2011dagger}.
On-policy distillation supervises student-generated states~\citep{chen2026futurebridge}, while OPSD uses privileged context to provide dense self-distillation targets~\citep{agarwal2024opd,gu2024minillm,zhao2026selfdistilled,jin2026eopd,li2026rethinking}.
Autoregressive distillation includes sequence-level supervision from teacher-generated outputs~\citep{kim2016seqkd} and objectives for efficient reuse of student-generated outputs~\citep{ko2024distillm}, while recent work directly transfers behavior from privileged-information-conditioned language-model teachers~\citep{penaloza2026privileged}.
Recent vision-language OPD variants decompose language and visual gradients or project teacher corrections onto locally realizable visual directions~\citep{yoon2026decomposed,xue2026fpopd}.
Multimodal extensions transfer reasoning across modalities or instantiate the privilege as localized crops, recoverable visual cues, or generated visual-thought traces~\citep{bousselham2025vold,yuan2026visionopd,tian2026vicur,li2026visualopsd}.

\paragraph{Reflection in language and multimodal reasoning.}
Reflection has been elicited through self-feedback and reinforcement learning in language models~\citep{madaan2023selfrefine,shinn2023reflexion,deepseekai2025r1}, and through reflection-aware reinforcement learning or iterative visual verification in multimodal models~\citep{zhou2025visualthinker,wan2025srpo,zhang2026mirror}.
Earlier approaches use tool-interactive critique, execution feedback, or backward verification to revise model outputs~\citep{gou2024critic,chen2024selfdebug,weng2023selfverification}.
Intrinsic self-correction can fail without reliable feedback, as controlled studies and reviews show~\citep{huang2024cannot,kamoi2024survey}.
Multimodal self-training further uses reflected rationales or vision-aware resampling to learn from failed trajectories~\citep{cheng2025r3v,zhong2026vista}, while broad benchmarking shows that the benefit of self-correction varies across tasks and correction strategies~\citep{tie2025correctbench}.
In our setting, reflection words are not explicitly supervised. They mark self-interruptions that become more likely when the reconstructed target suppresses an image-inconsistent continuation, followed by renewed visual reliance in subsequent tokens.

\section{Conclusion}
\label{sec:conclusion}

Privileged on-policy distillation suffers a supervision collapse when erroneous student prefixes overwhelm the teacher's visual evidence. We find that a token-specific visual corrective preference survives this collapse. OPD-Aha isolates this signal through an intra-teacher contrast and reconstructs a distillation target that suppresses image-inconsistent continuations. The trained student learns to interrupt flawed reasoning with reflection tokens, followed by renewed visual reliance after a short delay. OPD-Aha achieves consistent improvements across fine-grained perception and complex multimodal reasoning benchmarks. These gains support the value of separating visual preference from confounded language priors to preserve corrective supervision under flawed student prefixes.
\setlength{\bibsep}{0pt plus 0.3ex}
\bibliographystyle{opd_aha}
\bibliography{revision_opd}

@inproceedings{agarwal2024opd,
  author       = {Rishabh Agarwal and
                  Nino Vieillard and
                  Yongchao Zhou and
                  Piotr Stanczyk and
                  Sabela Ramos Garea and
                  Matthieu Geist and
                  Olivier Bachem},
  title        = {On-Policy Distillation of Language Models: Learning from Self-Generated
                  Mistakes},
  booktitle    = {The Twelfth International Conference on Learning Representations,
                  {ICLR} 2024, Vienna, Austria, May 7-11, 2024},
  publisher    = {OpenReview.net},
  year         = {2024},
  url          = {https://openreview.net/forum?id=3zKtaqxLhW},
  bibsource    = {dblp computer science bibliography, https://dblp.org}
}

@article{shao2024deepseekmath,
  author       = {Zhihong Shao and
                  Peiyi Wang and
                  Qihao Zhu and
                  Runxin Xu and
                  Junxiao Song and
                  Mingchuan Zhang and
                  Y. K. Li and
                  Y. Wu and
                  Daya Guo},
  title        = {DeepSeekMath: Pushing the Limits of Mathematical Reasoning in Open
                  Language Models},
  journal      = {CoRR},
  volume       = {abs/2402.03300},
  year         = {2024},
  url          = {https://doi.org/10.48550/arXiv.2402.03300},
  doi          = {10.48550/ARXIV.2402.03300},
  eprinttype   = {arXiv},
  eprint       = {2402.03300},
  bibsource    = {dblp computer science bibliography, https://dblp.org}
}

@article{zhao2026selfdistilled,
  author       = {Siyan Zhao and
                  Zhihui Xie and
                  Mengchen Liu and
                  Jing Huang and
                  Guan Pang and
                  Feiyu Chen and
                  Aditya Grover},
  title        = {Self-Distilled Reasoner: On-Policy Self-Distillation for Large Language
                  Models},
  journal      = {CoRR},
  volume       = {abs/2601.18734},
  year         = {2026},
  url          = {https://doi.org/10.48550/arXiv.2601.18734},
  doi          = {10.48550/ARXIV.2601.18734},
  eprinttype   = {arXiv},
  eprint       = {2601.18734},
  bibsource    = {dblp computer science bibliography, https://dblp.org}
}

@article{yuan2026visionopd,
  author       = {Qianhao Yuan and
                  Jie Lou and
                  Xing Yu and
                  Hongyu Lin and
                  Le Sun and
                  Xianpei Han and
                  Yaojie Lu},
  title        = {Vision-OPD: Learning to See Fine Details for Multimodal LLMs via On-Policy
                  Self-Distillation},
  journal      = {CoRR},
  volume       = {abs/2605.18740},
  year         = {2026},
  url          = {https://doi.org/10.48550/arXiv.2605.18740},
  doi          = {10.48550/ARXIV.2605.18740},
  eprinttype   = {arXiv},
  eprint       = {2605.18740},
  bibsource    = {dblp computer science bibliography, https://dblp.org}
}

@article{tian2026vicur,
  author       = {Kanghui Tian and
                  Siyuan Liu and
                  Ziang Yan and
                  Sheng Xia and
                  Shuai Dong and
                  Yi Wang},
  title        = {ViCuR: Visual Cues as Recoverable Privilege for Multimodal On-Policy
                  Distillation},
  journal      = {CoRR},
  volume       = {abs/2606.05718},
  year         = {2026},
  url          = {https://doi.org/10.48550/arXiv.2606.05718},
  doi          = {10.48550/ARXIV.2606.05718},
  eprinttype   = {arXiv},
  eprint       = {2606.05718},
  bibsource    = {dblp computer science bibliography, https://dblp.org}
}

@article{li2026videoopd,
  author       = {Jiaze Li and
                  Hao Yin and
                  Haoran Xu and
                  Boshen Xu and
                  Wenhui Tan and
                  Zewen He and
                  Jianzhong Ju and
                  Zhenbo Luo and
                  Jian Luan},
  title        = {Video-OPD: Efficient Post-Training of Multimodal Large Language Models
                  for Temporal Video Grounding via On-Policy Distillation},
  journal      = {CoRR},
  volume       = {abs/2602.02994},
  year         = {2026},
  url          = {https://doi.org/10.48550/arXiv.2602.02994},
  doi          = {10.48550/ARXIV.2602.02994},
  eprinttype   = {arXiv},
  eprint       = {2602.02994},
  bibsource    = {dblp computer science bibliography, https://dblp.org}
}

@article{jiang2026trd,
  author       = {Li Jiang and
                  Haoran Xu and
                  Yichuan Ding and
                  Amy Zhang},
  title        = {Trajectory-Refined Distillation},
  journal      = {CoRR},
  volume       = {abs/2606.08432},
  year         = {2026},
  url          = {https://doi.org/10.48550/arXiv.2606.08432},
  doi          = {10.48550/ARXIV.2606.08432},
  eprinttype   = {arXiv},
  eprint       = {2606.08432},
  bibsource    = {dblp computer science bibliography, https://dblp.org}
}

@article{xu2026relay,
  author       = {Haolei Xu and
                  Xiaowen Xu and
                  Haiwen Hong and
                  Zixuan Ni and
                  Hongxing Li and
                  Yiwen Qiu and
                  Weiming Lu and
                  Yongliang Shen},
  title        = {Pass the Baton: Trajectory-Relayed On-Policy Distillation},
  journal      = {CoRR},
  volume       = {abs/2607.26057},
  year         = {2026},
  url          = {https://doi.org/10.48550/arXiv.2607.26057},
  doi          = {10.48550/ARXIV.2607.26057},
  eprinttype   = {arXiv},
  eprint       = {2607.26057},
  bibsource    = {dblp computer science bibliography, https://dblp.org}
}

@article{chen2026futurebridge,
  author       = {Chishui Chen and
                  Yaoyou Fan and
                  Te Sun and
                  Yi Yang and
                  Chenghao Sun and
                  Delin Mao and
                  Hongbo Qiao and
                  Zuowei Zhang and
                  Junxi Wang and
                  Chenxing Sun and
                  Yangen Hu and
                  Lu Pan and
                  Xuyang Liu and
                  Linfeng Zhang},
  title        = {Look Ahead Before You Distill: Future Trajectory Validation of Teacher
                  Guidance for Agentic On-Policy Distillation},
  journal      = {CoRR},
  volume       = {abs/2608.01953},
  year         = {2026},
  url          = {https://doi.org/10.48550/arXiv.2608.01953},
  doi          = {10.48550/ARXIV.2608.01953},
  eprinttype   = {arXiv},
  eprint       = {2608.01953},
  bibsource    = {dblp computer science bibliography, https://dblp.org}
}

@article{li2026visualopsd,
  author       = {Pengyu Li and
                  Zhitao Gao and
                  Lingling Zhang and
                  Muye Huang and
                  Yuanming Li and
                  Fangzhi Xu and
                  Jun Liu},
  title        = {Visual-OPSD: Cross-Modal On-Policy Self-Distillation for Efficient
                  Unified Multimodal Reasoning},
  journal      = {CoRR},
  volume       = {abs/2606.18974},
  year         = {2026},
  url          = {https://doi.org/10.48550/arXiv.2606.18974},
  doi          = {10.48550/ARXIV.2606.18974},
  eprinttype   = {arXiv},
  eprint       = {2606.18974},
  bibsource    = {dblp computer science bibliography, https://dblp.org}
}

@inproceedings{leng2024vcd,
  author       = {Sicong Leng and
                  Hang Zhang and
                  Guanzheng Chen and
                  Xin Li and
                  Shijian Lu and
                  Chunyan Miao and
                  Lidong Bing},
  title        = {Mitigating Object Hallucinations in Large Vision-Language Models through
                  Visual Contrastive Decoding},
  booktitle    = {{IEEE/CVF} Conference on Computer Vision and Pattern Recognition,
                  {CVPR} 2024, Seattle, WA, USA, June 16-22, 2024},
  pages        = {13872--13882},
  publisher    = {{IEEE}},
  year         = {2024},
  url          = {https://doi.org/10.1109/CVPR52733.2024.01316},
  doi          = {10.1109/CVPR52733.2024.01316},
  bibsource    = {dblp computer science bibliography, https://dblp.org}
}

@article{madaan2023selfrefine,
  author       = {Aman Madaan and
                  Niket Tandon and
                  Prakhar Gupta and
                  Skyler Hallinan and
                  Luyu Gao and
                  Sarah Wiegreffe and
                  Uri Alon and
                  Nouha Dziri and
                  Shrimai Prabhumoye and
                  Yiming Yang and
                  Sean Welleck and
                  Bodhisattwa Prasad Majumder and
                  Shashank Gupta and
                  Amir Yazdanbakhsh and
                  Peter Clark},
  title        = {Self-Refine: Iterative Refinement with Self-Feedback},
  journal      = {CoRR},
  volume       = {abs/2303.17651},
  year         = {2023},
  url          = {https://doi.org/10.48550/arXiv.2303.17651},
  doi          = {10.48550/ARXIV.2303.17651},
  eprinttype   = {arXiv},
  eprint       = {2303.17651},
  bibsource    = {dblp computer science bibliography, https://dblp.org}
}

@inproceedings{shinn2023reflexion,
  author       = {Noah Shinn and
                  Federico Cassano and
                  Ashwin Gopinath and
                  Karthik Narasimhan and
                  Shunyu Yao},
  editor       = {Alice Oh and
                  Tristan Naumann and
                  Amir Globerson and
                  Kate Saenko and
                  Moritz Hardt and
                  Sergey Levine},
  title        = {Reflexion: language agents with verbal reinforcement learning},
  booktitle    = {Advances in Neural Information Processing Systems 36: Annual Conference
                  on Neural Information Processing Systems 2023, NeurIPS 2023, New Orleans,
                  LA, USA, December 10 - 16, 2023},
  year         = {2023},
  url          = {http://papers.nips.cc/paper\_files/paper/2023/hash/1b44b878bb782e6954cd888628510e90-Abstract-Conference.html},
  bibsource    = {dblp computer science bibliography, https://dblp.org}
}

@article{deepseekai2025r1,
  author       = {Daya Guo and
                  Dejian Yang and
                  Haowei Zhang and
                  Junxiao Song and
                  Peiyi Wang and
                  Qihao Zhu and
                  Runxin Xu and
                  Ruoyu Zhang and
                  Shirong Ma and
                  Xiao Bi and
                  Xiaokang Zhang and
                  Xingkai Yu and
                  Yu Wu and
                  Z. F. Wu and
                  Zhibin Gou and
                  Zhihong Shao and
                  Zhuoshu Li and
                  Ziyi Gao and
                  Aixin Liu and
                  Bing Xue and
                  Bingxuan Wang and
                  Bochao Wu and
                  Bei Feng and
                  Chengda Lu and
                  Chenggang Zhao and
                  Chengqi Deng and
                  Chong Ruan and
                  Damai Dai and
                  Deli Chen and
                  Dongjie Ji and
                  Erhang Li and
                  Fangyun Lin and
                  Fucong Dai and
                  Fuli Luo and
                  Guangbo Hao and
                  Guanting Chen and
                  Guowei Li and
                  Hao Zhang and
                  Hanwei Xu and
                  Honghui Ding and
                  Huazuo Gao and
                  Hui Qu and
                  Hui Li and
                  Jianzhong Guo and
                  Jiashi Li and
                  Jingchang Chen and
                  Jingyang Yuan and
                  Jinhao Tu and
                  Junjie Qiu and
                  Junlong Li and
                  J. L. Cai and
                  Jiaqi Ni and
                  Jian Liang and
                  Jin Chen and
                  Kai Dong and
                  Kai Hu and
                  Kaichao You and
                  Kaige Gao and
                  Kang Guan and
                  Kexin Huang and
                  Kuai Yu and
                  Lean Wang and
                  Lecong Zhang and
                  Liang Zhao and
                  Litong Wang and
                  Liyue Zhang and
                  Lei Xu and
                  Leyi Xia and
                  Mingchuan Zhang and
                  Minghua Zhang and
                  Minghui Tang and
                  Mingxu Zhou and
                  Meng Li and
                  Miaojun Wang and
                  Mingming Li and
                  Ning Tian and
                  Panpan Huang and
                  Peng Zhang and
                  Qiancheng Wang and
                  Qinyu Chen and
                  Qiushi Du and
                  Ruiqi Ge and
                  Ruisong Zhang and
                  Ruizhe Pan and
                  Runji Wang and
                  R. J. Chen and
                  R. L. Jin and
                  Ruyi Chen and
                  Shanghao Lu and
                  Shangyan Zhou and
                  Shanhuang Chen and
                  Shengfeng Ye and
                  Shiyu Wang and
                  Shuiping Yu and
                  Shunfeng Zhou and
                  Shuting Pan and
                  S. S. Li and
                  Shuang Zhou and
                  Shaoqing Wu and
                  Tao Yun and
                  Tian Pei and
                  Tianyu Sun and
                  Tao Wang and
                  Wangding Zeng and
                  Wen Liu and
                  Wenfeng Liang and
                  Wenjun Gao and
                  Wenqin Yu and
                  Wentao Zhang and
                  W. L. Xiao and
                  Wei An and
                  Xiaodong Liu and
                  Xiaohan Wang and
                  Xiaokang Chen and
                  Xiaotao Nie and
                  Xin Cheng and
                  Xin Liu and
                  Xin Xie and
                  Xingchao Liu and
                  Xinyu Yang and
                  Xinyuan Li and
                  Xuecheng Su and
                  Xuheng Lin and
                  X. Q. Li and
                  Xiangyue Jin and
                  Xiaojin Shen and
                  Xiaosha Chen and
                  Xiaowen Sun and
                  Xiaoxiang Wang and
                  Xinnan Song and
                  Xinyi Zhou and
                  Xianzu Wang and
                  Xinxia Shan and
                  Y. K. Li and
                  Y. Q. Wang and
                  Y. X. Wei and
                  Yang Zhang and
                  Yanhong Xu and
                  Yao Li and
                  Yao Zhao and
                  Yaofeng Sun and
                  Yaohui Wang and
                  Yi Yu and
                  Yichao Zhang and
                  Yifan Shi and
                  Yiliang Xiong and
                  Ying He and
                  Yishi Piao and
                  Yisong Wang and
                  Yixuan Tan and
                  Yiyang Ma and
                  Yiyuan Liu and
                  Yongqiang Guo and
                  Yuan Ou and
                  Yuduan Wang and
                  Yue Gong and
                  Yuheng Zou and
                  Yujia He and
                  Yunfan Xiong and
                  Yuxiang Luo and
                  Yuxiang You and
                  Yuxuan Liu and
                  Yuyang Zhou and
                  Y. X. Zhu and
                  Yanping Huang and
                  Yaohui Li and
                  Yi Zheng and
                  Yuchen Zhu and
                  Yunxian Ma and
                  Ying Tang and
                  Yukun Zha and
                  Yuting Yan and
                  Z. Z. Ren and
                  Zehui Ren and
                  Zhangli Sha and
                  Zhe Fu and
                  Zhean Xu and
                  Zhenda Xie and
                  Zhengyan Zhang and
                  Zhewen Hao and
                  Zhicheng Ma and
                  Zhigang Yan and
                  Zhiyu Wu and
                  Zihui Gu and
                  Zijia Zhu and
                  Zijun Liu and
                  Zilin Li and
                  Ziwei Xie and
                  Ziyang Song and
                  Zizheng Pan and
                  Zhen Huang and
                  Zhipeng Xu and
                  Zhongyu Zhang and
                  Zhen Zhang},
  title        = {DeepSeek-R1 incentivizes reasoning in LLMs through reinforcement learning},
  journal      = {Nat.},
  volume       = {645},
  number       = {8081},
  pages        = {633--638},
  year         = {2025},
  url          = {https://doi.org/10.1038/s41586-025-09422-z},
  doi          = {10.1038/S41586-025-09422-Z},
  bibsource    = {dblp computer science bibliography, https://dblp.org}
}

@article{zhou2025visualthinker,
  author       = {Hengguang Zhou and
                  Xirui Li and
                  Ruochen Wang and
                  Minhao Cheng and
                  Tianyi Zhou and
                  Cho{-}Jui Hsieh},
  title        = {R1-Zero's "Aha Moment" in Visual Reasoning on a 2B Non-SFT Model},
  journal      = {CoRR},
  volume       = {abs/2503.05132},
  year         = {2025},
  url          = {https://doi.org/10.48550/arXiv.2503.05132},
  doi          = {10.48550/ARXIV.2503.05132},
  eprinttype   = {arXiv},
  eprint       = {2503.05132},
  bibsource    = {dblp computer science bibliography, https://dblp.org}
}

@article{wan2025srpo,
  author       = {Zhongwei Wan and
                  Zhihao Dou and
                  Che Liu and
                  Yu Zhang and
                  Dongfei Cui and
                  Qinjian Zhao and
                  Hui Shen and
                  Jing Xiong and
                  Yi Xin and
                  Yifan Jiang and
                  Chaofan Tao and
                  Yangfan He and
                  Mi Zhang and
                  Shen Yan},
  title        = {{SRPO:} Enhancing Multimodal {LLM} Reasoning via Reflection-Aware
                  Reinforcement Learning},
  journal      = {CoRR},
  volume       = {abs/2506.01713},
  year         = {2025},
  url          = {https://doi.org/10.48550/arXiv.2506.01713},
  doi          = {10.48550/ARXIV.2506.01713},
  eprinttype   = {arXiv},
  eprint       = {2506.01713},
  bibsource    = {dblp computer science bibliography, https://dblp.org}
}

@article{zhang2026mirror,
  author       = {Haoyu Zhang and
                  Yuwei Wu and
                  Pengxiang Li and
                  Xintong Zhang and
                  Zhi Gao and
                  Rui Gao and
                  Mingyang Gao and
                  Che Sun and
                  Yunde Jia},
  title        = {{MIRROR:} Multimodal Iterative Reasoning via Reflection on Visual
                  Regions},
  journal      = {CoRR},
  volume       = {abs/2602.18746},
  year         = {2026},
  url          = {https://doi.org/10.48550/arXiv.2602.18746},
  doi          = {10.48550/ARXIV.2602.18746},
  eprinttype   = {arXiv},
  eprint       = {2602.18746},
  bibsource    = {dblp computer science bibliography, https://dblp.org}
}

@inproceedings{liu2023llava,
  author       = {Haotian Liu and
                  Chunyuan Li and
                  Qingyang Wu and
                  Yong Jae Lee},
  editor       = {Alice Oh and
                  Tristan Naumann and
                  Amir Globerson and
                  Kate Saenko and
                  Moritz Hardt and
                  Sergey Levine},
  title        = {Visual Instruction Tuning},
  booktitle    = {Advances in Neural Information Processing Systems 36: Annual Conference
                  on Neural Information Processing Systems 2023, NeurIPS 2023, New Orleans,
                  LA, USA, December 10 - 16, 2023},
  year         = {2023},
  url          = {http://papers.nips.cc/paper\_files/paper/2023/hash/6dcf277ea32ce3288914faf369fe6de0-Abstract-Conference.html},
  bibsource    = {dblp computer science bibliography, https://dblp.org}
}

@article{bai2023qwenvl,
  author       = {Jinze Bai and
                  Shuai Bai and
                  Shusheng Yang and
                  Shijie Wang and
                  Sinan Tan and
                  Peng Wang and
                  Junyang Lin and
                  Chang Zhou and
                  Jingren Zhou},
  title        = {Qwen-VL: {A} Frontier Large Vision-Language Model with Versatile Abilities},
  journal      = {CoRR},
  volume       = {abs/2308.12966},
  year         = {2023},
  url          = {https://doi.org/10.48550/arXiv.2308.12966},
  doi          = {10.48550/ARXIV.2308.12966},
  eprinttype   = {arXiv},
  eprint       = {2308.12966},
  bibsource    = {dblp computer science bibliography, https://dblp.org}
}

@inproceedings{li2023blip2,
  author       = {Junnan Li and
                  Dongxu Li and
                  Silvio Savarese and
                  Steven C. H. Hoi},
  editor       = {Andreas Krause and
                  Emma Brunskill and
                  Kyunghyun Cho and
                  Barbara Engelhardt and
                  Sivan Sabato and
                  Jonathan Scarlett},
  title        = {{BLIP-2:} Bootstrapping Language-Image Pre-training with Frozen Image
                  Encoders and Large Language Models},
  booktitle    = {International Conference on Machine Learning, {ICML} 2023, 23-29 July
                  2023, Honolulu, Hawaii, {USA}},
  series       = {Proceedings of Machine Learning Research},
  volume       = {202},
  pages        = {19730--19742},
  publisher    = {{PMLR}},
  year         = {2023},
  url          = {https://proceedings.mlr.press/v202/li23q.html},
  bibsource    = {dblp computer science bibliography, https://dblp.org}
}

@inproceedings{li2023pope,
  author       = {Yifan Li and
                  Yifan Du and
                  Kun Zhou and
                  Jinpeng Wang and
                  Wayne Xin Zhao and
                  Ji{-}Rong Wen},
  editor       = {Houda Bouamor and
                  Juan Pino and
                  Kalika Bali},
  title        = {Evaluating Object Hallucination in Large Vision-Language Models},
  booktitle    = {Proceedings of the 2023 Conference on Empirical Methods in Natural
                  Language Processing, {EMNLP} 2023, Singapore, December 6-10, 2023},
  pages        = {292--305},
  publisher    = {Association for Computational Linguistics},
  year         = {2023},
  url          = {https://doi.org/10.18653/v1/2023.emnlp-main.20},
  doi          = {10.18653/V1/2023.EMNLP-MAIN.20},
  bibsource    = {dblp computer science bibliography, https://dblp.org}
}

@inproceedings{favero2024m3id,
  author       = {Alessandro Favero and
                  Luca Zancato and
                  Matthew Trager and
                  Siddharth Choudhary and
                  Pramuditha Perera and
                  Alessandro Achille and
                  Ashwin Swaminathan and
                  Stefano Soatto},
  title        = {Multi-Modal Hallucination Control by Visual Information Grounding},
  booktitle    = {{IEEE/CVF} Conference on Computer Vision and Pattern Recognition,
                  {CVPR} 2024, Seattle, WA, USA, June 16-22, 2024},
  pages        = {14303--14312},
  publisher    = {{IEEE}},
  year         = {2024},
  url          = {https://doi.org/10.1109/CVPR52733.2024.01356},
  doi          = {10.1109/CVPR52733.2024.01356},
  bibsource    = {dblp computer science bibliography, https://dblp.org}
}

@article{guo2025lisa,
  author       = {Zhihui Guo and
                  Xin Man and
                  Hui Xu and
                  Jie Shao},
  title        = {{LISA:} {A} Layer-wise Integration and Suppression Approach for Hallucination
                  Mitigation in Multimodal Large Language Models},
  journal      = {CoRR},
  volume       = {abs/2507.19110},
  year         = {2025},
  url          = {https://doi.org/10.48550/arXiv.2507.19110},
  doi          = {10.48550/ARXIV.2507.19110},
  eprinttype   = {arXiv},
  eprint       = {2507.19110},
  bibsource    = {dblp computer science bibliography, https://dblp.org}
}

@article{li2026cognitive,
  author       = {Yinghui Li and
                  Jiayi Kuang and
                  Peng Xing and
                  Daixian Liu and
                  Junnan Dong and
                  Shu{-}Yu Guo and
                  Yangning Li and
                  Qingyu Zhou and
                  Wenhao Jiang and
                  Hai{-}Tao Zheng and
                  Ying Shen and
                  Liang Lin and
                  Philip S. Yu},
  title        = {Cognitive Mismatch in Multimodal Large Language Models for Discrete
                  Symbol Understanding},
  journal      = {CoRR},
  volume       = {abs/2603.18472},
  year         = {2026},
  url          = {https://doi.org/10.48550/arXiv.2603.18472},
  doi          = {10.48550/ARXIV.2603.18472},
  eprinttype   = {arXiv},
  eprint       = {2603.18472},
  bibsource    = {dblp computer science bibliography, https://dblp.org}
}

@inproceedings{he2025vhr,
  author       = {Jinghan He and
                  Kuan Zhu and
                  Haiyun Guo and
                  Junfeng Fang and
                  Zhenglin Hua and
                  Yuheng Jia and
                  Ming Tang and
                  Tat{-}Seng Chua and
                  Jinqiao Wang},
  editor       = {Wanxiang Che and
                  Joyce Nabende and
                  Ekaterina Shutova and
                  Mohammad Taher Pilehvar},
  title        = {Cracking the Code of Hallucination in LVLMs with Vision-aware Head
                  Divergence},
  booktitle    = {Proceedings of the 63rd Annual Meeting of the Association for Computational
                  Linguistics (Volume 1: Long Papers), {ACL} 2025, Vienna, Austria,
                  July 27 - August 1, 2025},
  pages        = {3488--3501},
  publisher    = {Association for Computational Linguistics},
  year         = {2025},
  url          = {https://doi.org/10.18653/v1/2025.acl-long.175},
  doi          = {10.18653/V1/2025.ACL-LONG.175},
  bibsource    = {dblp computer science bibliography, https://dblp.org}
}

@article{jin2026eopd,
  author       = {Woogyeol Jin and
                  Taywon Min and
                  Yongjin Yang and
                  Swanand Ravindra Kadhe and
                  Yi Zhou and
                  Dennis Wei and
                  Nathalie Baracaldo and
                  Kimin Lee},
  title        = {Entropy-Aware On-Policy Distillation of Language Models},
  journal      = {CoRR},
  volume       = {abs/2603.07079},
  year         = {2026},
  url          = {https://doi.org/10.48550/arXiv.2603.07079},
  doi          = {10.48550/ARXIV.2603.07079},
  eprinttype   = {arXiv},
  eprint       = {2603.07079},
  bibsource    = {dblp computer science bibliography, https://dblp.org}
}

@article{li2026rethinking,
  author       = {Yaxuan Li and
                  Yuxin Zuo and
                  Bingxiang He and
                  Jinqian Zhang and
                  Chaojun Xiao and
                  Cheng Qian and
                  Tianyu Yu and
                  Huan{-}ang Gao and
                  Wenkai Yang and
                  Zhiyuan Liu and
                  Ning Ding},
  title        = {Rethinking On-Policy Distillation of Large Language Models: Phenomenology,
                  Mechanism, and Recipe},
  journal      = {CoRR},
  volume       = {abs/2604.13016},
  year         = {2026},
  url          = {https://doi.org/10.48550/arXiv.2604.13016},
  doi          = {10.48550/ARXIV.2604.13016},
  eprinttype   = {arXiv},
  eprint       = {2604.13016},
  bibsource    = {dblp computer science bibliography, https://dblp.org}
}

@article{bousselham2025vold,
  author       = {Walid Bousselham and
                  Hilde Kuehne and
                  Cordelia Schmid},
  title        = {{VOLD:} Reasoning Transfer from LLMs to Vision-Language Models via
                  On-Policy Distillation},
  journal      = {CoRR},
  volume       = {abs/2510.23497},
  year         = {2025},
  url          = {https://doi.org/10.48550/arXiv.2510.23497},
  doi          = {10.48550/ARXIV.2510.23497},
  eprinttype   = {arXiv},
  eprint       = {2510.23497},
  bibsource    = {dblp computer science bibliography, https://dblp.org}
}

@inproceedings{gu2024minillm,
  author       = {Yuxian Gu and
                  Li Dong and
                  Furu Wei and
                  Minlie Huang},
  title        = {MiniLLM: Knowledge Distillation of Large Language Models},
  booktitle    = {The Twelfth International Conference on Learning Representations,
                  {ICLR} 2024, Vienna, Austria, May 7-11, 2024},
  publisher    = {OpenReview.net},
  year         = {2024},
  url          = {https://openreview.net/forum?id=5h0qf7IBZZ},
  bibsource    = {dblp computer science bibliography, https://dblp.org}
}

@article{yang2024pensieve,
  author       = {Dingchen Yang and
                  Bowen Cao and
                  Guang Chen and
                  Changjun Jiang},
  title        = {Pensieve: Retrospect-then-Compare Mitigates Visual Hallucination},
  journal      = {CoRR},
  volume       = {abs/2403.14401},
  year         = {2024},
  url          = {https://doi.org/10.48550/arXiv.2403.14401},
  doi          = {10.48550/ARXIV.2403.14401},
  eprinttype   = {arXiv},
  eprint       = {2403.14401},
  bibsource    = {dblp computer science bibliography, https://dblp.org}
}

@inproceedings{yin2025mirage,
  author       = {Hao Yin and
                  Guangzong Si and
                  Zilei Wang},
  editor       = {Danielle Belgrave and
                  Cheng Zhang and
                  Laura N. Montoya and
                  Hsuan{-}Tien Lin and
                  Razvan Pascanu and
                  Piotr Koniusz and
                  Marzyeh Ghassemi and
                  Nancy Chen and
                  Iv{\'{a}}n Vladimir Meza Ru{\'{\i}}z and
                  Arturo Loaiza{-}Bonilla},
  title        = {The Mirage of Performance Gains: Why Contrastive Decoding Fails to
                  Mitigate Object Hallucinations in MLLMs?},
  booktitle    = {Advances in Neural Information Processing Systems 38: Annual Conference
                  on Neural Information Processing Systems 2025, NeurIPS 2025, San Diego,
                  CA, USA, December 2-7, 2025 / Mexico City, Mexico, November 30 - December
                  5, 2025},
  year         = {2025},
  url          = {http://papers.nips.cc/paper\_files/paper/2025/hash/2f89a23a19d1617e7fb16d4f7a049ce2-Abstract-Conference.html},
  bibsource    = {dblp computer science bibliography, https://dblp.org}
}

@inproceedings{cheng2025r3v,
  author       = {Kanzhi Cheng and
                  Yantao Li and
                  Fangzhi Xu and
                  Jianbing Zhang and
                  Hao Zhou and
                  Yang Liu},
  editor       = {Luis Chiruzzo and
                  Alan Ritter and
                  Lu Wang},
  title        = {Vision-Language Models Can Self-Improve Reasoning via Reflection},
  booktitle    = {Proceedings of the 2025 Conference of the Nations of the Americas
                  Chapter of the Association for Computational Linguistics: Human Language
                  Technologies, {NAACL} 2025 - Volume 1: Long Papers, Albuquerque, New
                  Mexico, USA, April 29 - May 4, 2025},
  pages        = {8876--8892},
  publisher    = {Association for Computational Linguistics},
  year         = {2025},
  url          = {https://doi.org/10.18653/v1/2025.naacl-long.447},
  doi          = {10.18653/V1/2025.NAACL-LONG.447},
  bibsource    = {dblp computer science bibliography, https://dblp.org}
}

@article{zhong2026vista,
  author       = {Qihuang Zhong and
                  Liang Ding and
                  Wenjie Xuan and
                  Juhua Liu and
                  Bo Du and
                  Dacheng Tao},
  title        = {Learn to Think: Improving Multimodal Reasoning through Vision-Aware
                  Self-Improvement Training},
  journal      = {CoRR},
  volume       = {abs/2605.11931},
  year         = {2026},
  url          = {https://doi.org/10.48550/arXiv.2605.11931},
  doi          = {10.48550/ARXIV.2605.11931},
  eprinttype   = {arXiv},
  eprint       = {2605.11931},
  bibsource    = {dblp computer science bibliography, https://dblp.org}
}

@article{tie2025correctbench,
  author       = {Guiyao Tie and
                  Zenghui Yuan and
                  Zeli Zhao and
                  Chaoran Hu and
                  Tianhe Gu and
                  Ruihang Zhang and
                  Sizhe Zhang and
                  Junran Wu and
                  Xiaoyue Tu and
                  Ming Jin and
                  Qingsong Wen and
                  Lixing Chen and
                  Pan Zhou and
                  Lichao Sun},
  title        = {Can LLMs Correct Themselves? {A} Benchmark of Self-Correction in LLMs},
  journal      = {CoRR},
  volume       = {abs/2510.16062},
  year         = {2025},
  url          = {https://doi.org/10.48550/arXiv.2510.16062},
  doi          = {10.48550/ARXIV.2510.16062},
  eprinttype   = {arXiv},
  eprint       = {2510.16062},
  bibsource    = {dblp computer science bibliography, https://dblp.org}
}

@article{yoon2026decomposed,
  author       = {Hee Suk Yoon and
                  Eunseop Yoon and
                  Jaehyun Jang and
                  SooHwan Eom and
                  Ji Woo Hong and
                  Mark Hasegawa{-}Johnson and
                  Qi Dai and
                  Chong Luo and
                  Chang D. Yoo},
  title        = {Decomposed On-Policy Distillation for Vision-Language Reasoning: Steering
                  Gradients for Visual Grounding},
  journal      = {CoRR},
  volume       = {abs/2606.00564},
  year         = {2026},
  url          = {https://doi.org/10.48550/arXiv.2606.00564},
  doi          = {10.48550/ARXIV.2606.00564},
  eprinttype   = {arXiv},
  eprint       = {2606.00564},
  bibsource    = {dblp computer science bibliography, https://dblp.org}
}

@article{xue2026fpopd,
  author       = {Leyan Xue and
                  Feng Xiong and
                  Mingjun Ma and
                  Changqing Zhang},
  title        = {Distill What the Student Can See: Fisher-Projected On-Policy Distillation
                  for Vision-Language Models},
  journal      = {CoRR},
  volume       = {abs/2608.01263},
  year         = {2026},
  url          = {https://doi.org/10.48550/arXiv.2608.01263},
  doi          = {10.48550/ARXIV.2608.01263},
  eprinttype   = {arXiv},
  eprint       = {2608.01263},
  bibsource    = {dblp computer science bibliography, https://dblp.org}
}

@article{zhao2025cicd,
  author       = {Jianfei Zhao and
                  Feng Zhang and
                  Xin Sun and
                  Chong Feng},
  title        = {Cross-Image Contrastive Decoding: Precise, Lossless Suppression of
                  Language Priors in Large Vision-Language Models},
  journal      = {CoRR},
  volume       = {abs/2505.10634},
  year         = {2025},
  url          = {https://doi.org/10.48550/arXiv.2505.10634},
  doi          = {10.48550/ARXIV.2505.10634},
  eprinttype   = {arXiv},
  eprint       = {2505.10634},
  bibsource    = {dblp computer science bibliography, https://dblp.org}
}

@article{chen2026resdec,
  author       = {Xinrong Chen and
                  Xu Chu and
                  Yingmin Qiu and
                  Hengyuan Zhang and
                  Jing Xiong and
                  Shiyu Tang and
                  Shuai Liu and
                  Shaokang Yang and
                  Cheng Yang and
                  Hayden Kwok{-}Hay So and
                  Ngai Wong},
  title        = {Residual Decoding: Mitigating Hallucinations in Large Vision-Language
                  Models via History-Aware Residual Guidance},
  journal      = {CoRR},
  volume       = {abs/2602.01047},
  year         = {2026},
  url          = {https://doi.org/10.48550/arXiv.2602.01047},
  doi          = {10.48550/ARXIV.2602.01047},
  eprinttype   = {arXiv},
  eprint       = {2602.01047},
  bibsource    = {dblp computer science bibliography, https://dblp.org}
}

@article{hinton2015distilling,
  author       = {Geoffrey E. Hinton and
                  Oriol Vinyals and
                  Jeffrey Dean},
  title        = {Distilling the Knowledge in a Neural Network},
  journal      = {CoRR},
  volume       = {abs/1503.02531},
  year         = {2015},
  url          = {http://arxiv.org/abs/1503.02531},
  eprinttype   = {arXiv},
  eprint       = {1503.02531},
  bibsource    = {dblp computer science bibliography, https://dblp.org}
}

@article{vapnik2009privileged,
  author       = {Vladimir Vapnik and
                  Akshay Vashist},
  title        = {A new learning paradigm: Learning using privileged information},
  journal      = {Neural Networks},
  volume       = {22},
  number       = {5-6},
  pages        = {544--557},
  year         = {2009},
  url          = {https://doi.org/10.1016/j.neunet.2009.06.042},
  doi          = {10.1016/J.NEUNET.2009.06.042},
  bibsource    = {dblp computer science bibliography, https://dblp.org}
}

@inproceedings{lopezpaz2016unifying,
  author       = {David Lopez{-}Paz and
                  L{\'{e}}on Bottou and
                  Bernhard Sch{\"{o}}lkopf and
                  Vladimir Vapnik},
  editor       = {Yoshua Bengio and
                  Yann LeCun},
  title        = {Unifying distillation and privileged information},
  booktitle    = {4th International Conference on Learning Representations, {ICLR} 2016,
                  San Juan, Puerto Rico, May 2-4, 2016, Conference Track Proceedings},
  year         = {2016},
  url          = {http://arxiv.org/abs/1511.03643},
  bibsource    = {dblp computer science bibliography, https://dblp.org}
}

@inproceedings{ko2024distillm,
  author       = {Jongwoo Ko and
                  Sungnyun Kim and
                  Tianyi Chen and
                  Se{-}Young Yun},
  editor       = {Ruslan Salakhutdinov and
                  Zico Kolter and
                  Katherine A. Heller and
                  Adrian Weller and
                  Nuria Oliver and
                  Jonathan Scarlett and
                  Felix Berkenkamp},
  title        = {DistiLLM: Towards Streamlined Distillation for Large Language Models},
  booktitle    = {Forty-first International Conference on Machine Learning, {ICML} 2024,
                  Vienna, Austria, July 21-27, 2024},
  series       = {Proceedings of Machine Learning Research},
  volume       = {235},
  pages        = {24872--24895},
  publisher    = {{PMLR} / OpenReview.net},
  year         = {2024},
  url          = {https://proceedings.mlr.press/v235/ko24c.html},
  bibsource    = {dblp computer science bibliography, https://dblp.org}
}

@article{penaloza2026privileged,
  author       = {Emiliano Penaloza and
                  Dheeraj Vattikonda and
                  Nicolas Gontier and
                  Alexandre Lacoste and
                  Laurent Charlin and
                  Massimo Caccia},
  title        = {Privileged Information Distillation for Language Models},
  journal      = {CoRR},
  volume       = {abs/2602.04942},
  year         = {2026},
  url          = {https://doi.org/10.48550/arXiv.2602.04942},
  doi          = {10.48550/ARXIV.2602.04942},
  eprinttype   = {arXiv},
  eprint       = {2602.04942},
  bibsource    = {dblp computer science bibliography, https://dblp.org}
}

@inproceedings{gou2024critic,
  author       = {Zhibin Gou and
                  Zhihong Shao and
                  Yeyun Gong and
                  Yelong Shen and
                  Yujiu Yang and
                  Nan Duan and
                  Weizhu Chen},
  title        = {{CRITIC:} Large Language Models Can Self-Correct with Tool-Interactive
                  Critiquing},
  booktitle    = {The Twelfth International Conference on Learning Representations,
                  {ICLR} 2024, Vienna, Austria, May 7-11, 2024},
  publisher    = {OpenReview.net},
  year         = {2024},
  url          = {https://openreview.net/forum?id=Sx038qxjek},
  bibsource    = {dblp computer science bibliography, https://dblp.org}
}

@inproceedings{chen2024selfdebug,
  author       = {Xinyun Chen and
                  Maxwell Lin and
                  Nathanael Sch{\"{a}}rli and
                  Denny Zhou},
  title        = {Teaching Large Language Models to Self-Debug},
  booktitle    = {The Twelfth International Conference on Learning Representations,
                  {ICLR} 2024, Vienna, Austria, May 7-11, 2024},
  publisher    = {OpenReview.net},
  year         = {2024},
  url          = {https://openreview.net/forum?id=KuPixIqPiq},
  bibsource    = {dblp computer science bibliography, https://dblp.org}
}

@inproceedings{weng2023selfverification,
  author       = {Yixuan Weng and
                  Minjun Zhu and
                  Fei Xia and
                  Bin Li and
                  Shizhu He and
                  Shengping Liu and
                  Bin Sun and
                  Kang Liu and
                  Jun Zhao},
  editor       = {Houda Bouamor and
                  Juan Pino and
                  Kalika Bali},
  title        = {Large Language Models are Better Reasoners with Self-Verification},
  booktitle    = {Findings of the Association for Computational Linguistics: {EMNLP}
                  2023, Singapore, December 6-10, 2023},
  series       = {Findings of {ACL}},
  volume       = {{EMNLP} 2023},
  pages        = {2550--2575},
  publisher    = {Association for Computational Linguistics},
  year         = {2023},
  url          = {https://doi.org/10.18653/v1/2023.findings-emnlp.167},
  doi          = {10.18653/V1/2023.FINDINGS-EMNLP.167},
  bibsource    = {dblp computer science bibliography, https://dblp.org}
}

@inproceedings{huang2024cannot,
  author       = {Jie Huang and
                  Xinyun Chen and
                  Swaroop Mishra and
                  Huaixiu Steven Zheng and
                  Adams Wei Yu and
                  Xinying Song and
                  Denny Zhou},
  title        = {Large Language Models Cannot Self-Correct Reasoning Yet},
  booktitle    = {The Twelfth International Conference on Learning Representations,
                  {ICLR} 2024, Vienna, Austria, May 7-11, 2024},
  publisher    = {OpenReview.net},
  year         = {2024},
  url          = {https://openreview.net/forum?id=IkmD3fKBPQ},
  bibsource    = {dblp computer science bibliography, https://dblp.org}
}

@article{kamoi2024survey,
  author       = {Ryo Kamoi and
                  Yusen Zhang and
                  Nan Zhang and
                  Jiawei Han and
                  Rui Zhang},
  title        = {When Can LLMs \emph{Actually} Correct Their Own Mistakes? {A} Critical
                  Survey of Self-Correction of LLMs},
  journal      = {Trans. Assoc. Comput. Linguistics},
  volume       = {12},
  pages        = {1417--1440},
  year         = {2024},
  url          = {https://doi.org/10.1162/tacl\_a\_00713},
  doi          = {10.1162/TACL\_A\_00713},
  bibsource    = {dblp computer science bibliography, https://dblp.org}
}

@inproceedings{wu2024vstar,
  author       = {Penghao Wu and
                  Saining Xie},
  title        = {V*: Guided Visual Search as a Core Mechanism in Multimodal LLMs},
  booktitle    = {{IEEE/CVF} Conference on Computer Vision and Pattern Recognition,
                  {CVPR} 2024, Seattle, WA, USA, June 16-22, 2024},
  pages        = {13084--13094},
  publisher    = {{IEEE}},
  year         = {2024},
  url          = {https://doi.org/10.1109/CVPR52733.2024.01243},
  doi          = {10.1109/CVPR52733.2024.01243},
  bibsource    = {dblp computer science bibliography, https://dblp.org}
}

@article{wei2026zoombench,
  author       = {Lai Wei and
                  Liangbo He and
                  Jun Lan and
                  Lingzhong Dong and
                  Yutong Cai and
                  Siyuan Li and
                  Huijia Zhu and
                  Weiqiang Wang and
                  Linghe Kong and
                  Yue Wang and
                  Zhuosheng Zhang and
                  Weiran Huang},
  title        = {Zooming without Zooming: Region-to-Image Distillation for Fine-Grained
                  Multimodal Perception},
  journal      = {CoRR},
  volume       = {abs/2602.11858},
  year         = {2026},
  url          = {https://doi.org/10.48550/arXiv.2602.11858},
  doi          = {10.48550/ARXIV.2602.11858},
  eprinttype   = {arXiv},
  eprint       = {2602.11858},
  bibsource    = {dblp computer science bibliography, https://dblp.org}
}

@inproceedings{wang2025hrbench,
  author       = {Wenbin Wang and
                  Liang Ding and
                  Minyan Zeng and
                  Xiabin Zhou and
                  Li Shen and
                  Yong Luo and
                  Wei Yu and
                  Dacheng Tao},
  editor       = {Toby Walsh and
                  Julie Shah and
                  Zico Kolter},
  title        = {Divide, Conquer and Combine: {A} Training-Free Framework for High-Resolution
                  Image Perception in Multimodal Large Language Models},
  booktitle    = {Thirty-Ninth {AAAI} Conference on Artificial Intelligence, Thirty-Seventh
                  Conference on Innovative Applications of Artificial Intelligence,
                  Fifteenth Symposium on Educational Advances in Artificial Intelligence,
                  {AAAI} 2025, Philadelphia, PA, USA, February 25 - March 4, 2025},
  pages        = {7907--7915},
  publisher    = {{AAAI} Press},
  year         = {2025},
  url          = {https://doi.org/10.1609/aaai.v39i8.32852},
  doi          = {10.1609/AAAI.V39I8.32852},
  bibsource    = {dblp computer science bibliography, https://dblp.org}
}

@inproceedings{zhang2025mmerealworld,
  author       = {Yifan Zhang and
                  Huanyu Zhang and
                  Haochen Tian and
                  Chaoyou Fu and
                  Shuangqing Zhang and
                  Junfei Wu and
                  Feng Li and
                  Kun Wang and
                  Qingsong Wen and
                  Zhang Zhang and
                  Liang Wang and
                  Rong Jin},
  title        = {MME-RealWorld: Could Your Multimodal {LLM} Challenge High-Resolution
                  Real-World Scenarios that are Difficult for Humans?},
  booktitle    = {The Thirteenth International Conference on Learning Representations,
                  {ICLR} 2025, Singapore, April 24-28, 2025},
  publisher    = {OpenReview.net},
  year         = {2025},
  url          = {https://openreview.net/forum?id=k5VHHgsRbi},
  bibsource    = {dblp computer science bibliography, https://dblp.org}
}

@inproceedings{zhang2024mathverse,
  author       = {Renrui Zhang and
                  Dongzhi Jiang and
                  Yichi Zhang and
                  Haokun Lin and
                  Ziyu Guo and
                  Pengshuo Qiu and
                  Aojun Zhou and
                  Pan Lu and
                  Kai{-}Wei Chang and
                  Yu Qiao and
                  Peng Gao and
                  Hongsheng Li},
  editor       = {Ales Leonardis and
                  Elisa Ricci and
                  Stefan Roth and
                  Olga Russakovsky and
                  Torsten Sattler and
                  G{\"{u}}l Varol},
  title        = {{MATHVERSE:} Does Your Multi-modal {LLM} Truly See the Diagrams in
                  Visual Math Problems?},
  booktitle    = {Computer Vision - {ECCV} 2024 - 18th European Conference, Milan, Italy,
                  September 29-October 4, 2024, Proceedings, Part {VIII}},
  series       = {Lecture Notes in Computer Science},
  volume       = {15066},
  pages        = {169--186},
  publisher    = {Springer},
  year         = {2024},
  url          = {https://doi.org/10.1007/978-3-031-73242-3\_10},
  doi          = {10.1007/978-3-031-73242-3\_10},
  bibsource    = {dblp computer science bibliography, https://dblp.org}
}

@inproceedings{lu2024mathvista,
  author       = {Pan Lu and
                  Hritik Bansal and
                  Tony Xia and
                  Jiacheng Liu and
                  Chunyuan Li and
                  Hannaneh Hajishirzi and
                  Hao Cheng and
                  Kai{-}Wei Chang and
                  Michel Galley and
                  Jianfeng Gao},
  title        = {MathVista: Evaluating Mathematical Reasoning of Foundation Models
                  in Visual Contexts},
  booktitle    = {The Twelfth International Conference on Learning Representations,
                  {ICLR} 2024, Vienna, Austria, May 7-11, 2024},
  publisher    = {OpenReview.net},
  year         = {2024},
  url          = {https://openreview.net/forum?id=KUNzEQMWU7},
  bibsource    = {dblp computer science bibliography, https://dblp.org}
}

@inproceedings{wang2024mathvision,
  author       = {Ke Wang and
                  Junting Pan and
                  Weikang Shi and
                  Zimu Lu and
                  Houxing Ren and
                  Aojun Zhou and
                  Mingjie Zhan and
                  Hongsheng Li},
  editor       = {Amir Globersons and
                  Lester Mackey and
                  Danielle Belgrave and
                  Angela Fan and
                  Ulrich Paquet and
                  Jakub M. Tomczak and
                  Cheng Zhang},
  title        = {Measuring Multimodal Mathematical Reasoning with MATH-Vision Dataset},
  booktitle    = {Advances in Neural Information Processing Systems 37: Annual Conference
                  on Neural Information Processing Systems 2024, NeurIPS 2024, Vancouver,
                  BC, Canada, December 10 - 15, 2024},
  year         = {2024},
  url          = {http://papers.nips.cc/paper\_files/paper/2024/hash/ad0edc7d5fa1a783f063646968b7315b-Abstract-Datasets\_and\_Benchmarks\_Track.html},
  bibsource    = {dblp computer science bibliography, https://dblp.org}
}

@inproceedings{qiao2025wemath,
  author       = {Runqi Qiao and
                  Qiuna Tan and
                  Guanting Dong and
                  Minhui Wu and
                  Chong Sun and
                  Xiaoshuai Song and
                  Jiapeng Wang and
                  Zhuoma Gongque and
                  Shanglin Lei and
                  Yifan Zhang and
                  Zhe Wei and
                  Miaoxuan Zhang and
                  Runfeng Qiao and
                  Xiao Zong and
                  Yida Xu and
                  Peiqing Yang and
                  Zhimin Bao and
                  Muxi Diao and
                  Chen Li and
                  Honggang Zhang},
  editor       = {Wanxiang Che and
                  Joyce Nabende and
                  Ekaterina Shutova and
                  Mohammad Taher Pilehvar},
  title        = {We-Math: Does Your Large Multimodal Model Achieve Human-like Mathematical
                  Reasoning?},
  booktitle    = {Proceedings of the 63rd Annual Meeting of the Association for Computational
                  Linguistics (Volume 1: Long Papers), {ACL} 2025, Vienna, Austria,
                  July 27 - August 1, 2025},
  pages        = {20023--20070},
  publisher    = {Association for Computational Linguistics},
  year         = {2025},
  url          = {https://doi.org/10.18653/v1/2025.acl-long.983},
  doi          = {10.18653/V1/2025.ACL-LONG.983},
  bibsource    = {dblp computer science bibliography, https://dblp.org}
}

@inproceedings{zou2025dynamath,
  author       = {Chengke Zou and
                  Xingang Guo and
                  Rui Yang and
                  Junyu Zhang and
                  Bin Hu and
                  Huan Zhang},
  title        = {DynaMath: {A} Dynamic Visual Benchmark for Evaluating Mathematical
                  Reasoning Robustness of Vision Language Models},
  booktitle    = {The Thirteenth International Conference on Learning Representations,
                  {ICLR} 2025, Singapore, April 24-28, 2025},
  publisher    = {OpenReview.net},
  year         = {2025},
  url          = {https://openreview.net/forum?id=VOAMTA8jKu},
  bibsource    = {dblp computer science bibliography, https://dblp.org}
}

@inproceedings{ross2011dagger,
  author       = {St{\'{e}}phane Ross and
                  Geoffrey J. Gordon and
                  Drew Bagnell},
  editor       = {Geoffrey J. Gordon and
                  David B. Dunson and
                  Miroslav Dud{\'{\i}}k},
  title        = {A Reduction of Imitation Learning and Structured Prediction to No-Regret
                  Online Learning},
  booktitle    = {Proceedings of the Fourteenth International Conference on Artificial
                  Intelligence and Statistics, {AISTATS} 2011, Fort Lauderdale, USA,
                  April 11-13, 2011},
  series       = {{JMLR} Proceedings},
  volume       = {15},
  pages        = {627--635},
  publisher    = {JMLR.org},
  year         = {2011},
  url          = {http://proceedings.mlr.press/v15/ross11a/ross11a.pdf},
  bibsource    = {dblp computer science bibliography, https://dblp.org}
}

@inproceedings{kim2016seqkd,
  author       = {Yoon Kim and
                  Alexander M. Rush},
  editor       = {Jian Su and
                  Xavier Carreras and
                  Kevin Duh},
  title        = {Sequence-Level Knowledge Distillation},
  booktitle    = {Proceedings of the 2016 Conference on Empirical Methods in Natural
                  Language Processing, {EMNLP} 2016, Austin, Texas, USA, November 1-4,
                  2016},
  pages        = {1317--1327},
  publisher    = {The Association for Computational Linguistics},
  year         = {2016},
  url          = {https://doi.org/10.18653/v1/d16-1139},
  doi          = {10.18653/V1/D16-1139},
  bibsource    = {dblp computer science bibliography, https://dblp.org}
}

\clearpage
\appendix
\section{Theoretical Properties of Target Reconstruction}
\label{sec:target_optimality_appendix}

The reconstructed target $q_t$ uniquely maximizes the KL-regularized objective in Equation~\ref{eq:evidence_regularized_target}. To make its normalization and global optimality explicit, assume that the teacher softmax assigns positive probability over the vocabulary and introduce a multiplier $\lambda$ for the simplex constraint:
\begin{equation}
\begin{aligned}
    \mathcal J(q,\lambda)
    ={}&
    \beta\sum_{v\in\mathcal V}q(v)u_t(v)
    -
    \sum_{v\in\mathcal V}q(v)
    \log\frac{q(v)}{p_t^{+}(v)} \\
    &+
    \lambda\left(\sum_{v\in\mathcal V}q(v)-1\right).
\end{aligned}
\label{eq:target_lagrangian}
\end{equation}
At the optimum, each token balances its visual preference against its log-probability shift from the privileged teacher:
\begin{equation}
    \frac{\partial\mathcal J}{\partial q(v)}
    =
    \beta u_t(v)
    -
    \log\frac{q(v)}{p_t^{+}(v)}
    -1+\lambda
    =0,
\label{eq:target_lagrangian_stationarity}
\end{equation}
This balance yields $q(v)\propto p_t^{+}(v)\exp(\beta u_t(v))$. The negative KL term is strictly concave on this support, while the expected visual alignment is linear in $q$. The normalized distribution in Equation~\ref{eq:counterfactual_target} is therefore the unique global optimum.

Target reconstruction reverses an erroneous teacher preference when the surviving visual separation is sufficiently strong. Let $v^{+}$ denote a token supporting the correct answer and $v^{-}$ a token favored by the accumulated hallucination, with $u_t(v^{+})>u_t(v^{-})$. Even when the privileged teacher distribution favors $v^{-}$, the reconstructed target prioritizes $v^{+}$ precisely when
\begin{equation}
    \beta
    >
    \frac{
        \log p_t^{+}(v^{-})
        -
        \log p_t^{+}(v^{+})
    }{
        u_t(v^{+})
        -
        u_t(v^{-})
    }.
\label{eq:minimum_beta_for_correction}
\end{equation}
This threshold separates two regimes. When $p_t^{+}$ still favors $v^{+}$, the numerator is nonpositive and no positive minimum reconstruction strength is required. Once the accumulated student prefix shifts the teacher toward $v^{-}$, the threshold increases with the erroneous base preference and decreases with the surviving visual separation. Later prefix states therefore require stronger reconstruction whenever linguistic momentum intensifies or the visual separation weakens.

The reconstruction strength also determines the admissible departure from the privileged target. For every $\beta>0$, define $\varepsilon_{\beta}=D_{\mathrm{KL}}(q_t\,\|\,p_t^{+})$. The reconstructed target solves
\begin{equation}
    q_t
    =
    \underset{q\in\Delta(\mathcal V)}{\arg\max}
    \ \mathbb E_{v\sim q}\!\left[u_t(v)\right]
    \quad
    \text{subject to}
    \quad
    D_{\mathrm{KL}}\!\left(q\,\|\,p_t^{+}\right)
    \leq
    \varepsilon_{\beta}.
\label{eq:constrained_target_problem}
\end{equation}
Within this KL neighborhood, $q_t$ attains the strongest expected alignment with the visual preference. The privileged teacher supplies the reference distribution, the real--null difference determines the direction of movement, and $\beta$ selects how far the target travels along that direction.

Across the resulting one-parameter family, stronger reconstruction monotonically increases both expected visual preference and departure from the privileged target. Let $q_{t,\beta}$ denote the reconstructed target at strength $\beta$:
\begin{equation}
\begin{aligned}
    \frac{\mathrm d}{\mathrm d\beta}
    \mathbb E_{v\sim q_{t,\beta}}[u_t(v)]
    &=
    \operatorname{Var}_{v\sim q_{t,\beta}}[u_t(v)]
    \geq 0, \\
    \frac{\mathrm d}{\mathrm d\beta}
    D_{\mathrm{KL}}(q_{t,\beta}\,\|\,p_t^{+})
    &=
    \beta\operatorname{Var}_{v\sim q_{t,\beta}}[u_t(v)]
    \geq 0.
\end{aligned}
\label{eq:beta_monotonicity}
\end{equation}
Expected visual alignment and KL departure therefore vary monotonically with $\beta$. The reconstruction strength acts as a direct control parameter for how aggressively the target moves away from the privileged teacher distribution along the visual-preference direction.

\clearpage
\section{Extended Analyses of the Correction Mechanism}

\subsection{Learning When to Reflect}
\label{sec:reflection_dynamics_appendix}

To understand whether the student learns to interrupt flawed reasoning or merely adopts a broader bias toward reflection vocabulary, we track its preference for reflection tokens across intermediate training checkpoints. We compare the student's predictions on trajectories that ultimately lead to reflection against comparable non-reflection trajectories. This contrast isolates the underlying learning dynamic and reveals how the student internalizes the state-specific correction provided by the reconstructed target.

\begin{figure}[!htbp]
    \centering
    \includegraphics[width=\textwidth]{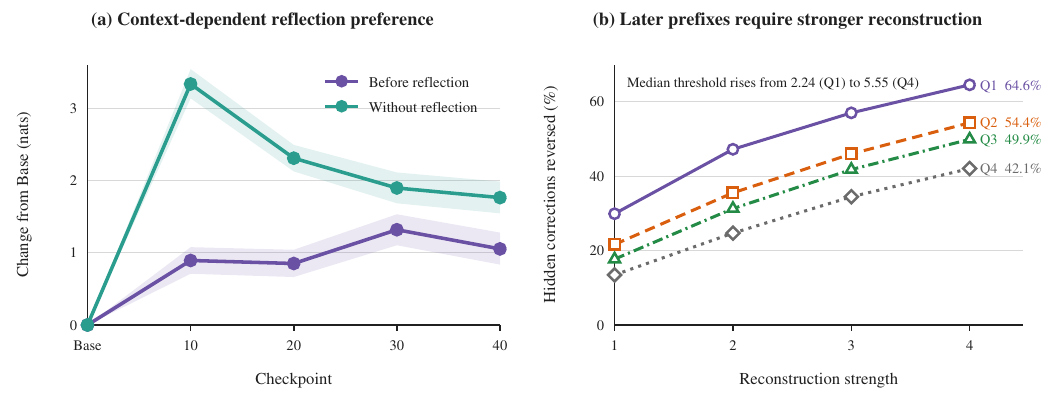}
    \caption{Context-dependent reflection and late-prefix target recovery. \textbf{(a)} During training, the increase in reflection-token preference recedes at positions without reflection but remains elevated immediately before reflection. \textbf{(b)} Stronger reconstruction reverses more wrong teacher preferences, although later prefix quartiles require greater strength.}
    \label{fig:reflection_checkpoint_dynamics}
\end{figure}

We observe a distinct two-stage behavioral shift during optimization. Early in training, the student raises the probability of reflection tokens across all evaluated positions. As training progresses, this broad increase diminishes at non-reflection states while remaining strongly elevated precisely at the states immediately preceding a reflection token (Figure~\ref{fig:reflection_checkpoint_dynamics}a). The trained student therefore moves beyond a simple vocabulary shift. It learns to recognize the specific states where the current explanation conflicts with visual evidence and uses reflection to interrupt that explanation.

\FloatBarrier
\Needspace{8\baselineskip}

\subsection{Reconstructed Supervision Drives Evidence-Specific Recovery}
\label{sec:hidden_correction_appendix}

\paragraph{Evidence-specific target recovery.}
To confirm that downstream corrections arise from the structured visual preference rather than a generic change in target magnitude or uncertainty, we isolate the token-level and spatial dependencies of the reconstructed target. We compare the true reconstructed target against alternatives that shuffle the token identities while preserving the magnitude or uncertainty of the target change. At the spatial level, we replace the task-relevant evidence crop with non-overlapping visual content from the same image to determine whether the correction relies on the designated evidence region.

\begin{figure}[!htbp]
    \centering
    \includegraphics[width=\textwidth]{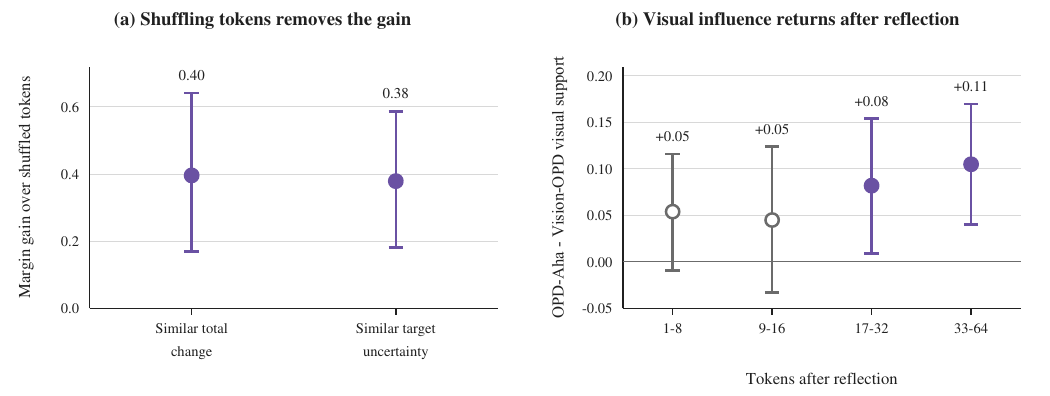}
    \caption{Token-specific correction and renewed visual support. \textbf{(a)} The visual preference signal produces a larger correct-option margin than token-shuffled alternatives with comparable target change or uncertainty. \textbf{(b)} After reflection, OPD-Aha gains more visual support for its generated tokens than Vision-OPD.}
    \label{fig:hidden_correction_mechanism}
\end{figure}

\begin{figure}[!htbp]
    \centering
    \includegraphics[width=0.98\textwidth]{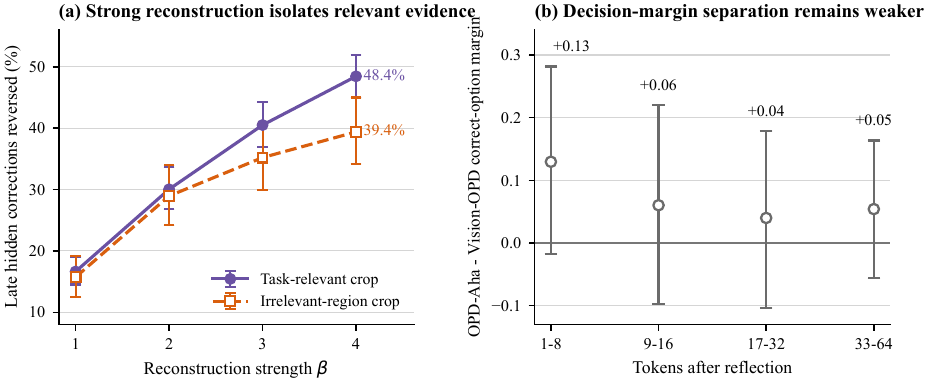}
    \caption{Evidence specificity and answer-margin dynamics after reflection. \textbf{(a)} As reconstruction strengthens, the task-relevant crop reverses more late wrong preferences than a same-shaped, non-overlapping crop from the same image. \textbf{(b)} At the same reflection prefixes, OPD-Aha maintains a larger correct-option margin than Vision-OPD.}
    \label{fig:evidence_specificity_decision}
\end{figure}

We find that target recovery depends on the precise corrective direction and the relevant visual evidence. Stronger reconstruction overturns more erroneous teacher preferences and reaches later states under accumulated hallucinations (Figure~\ref{fig:reflection_checkpoint_dynamics}b). Shuffling the visual preference removes most of the correct-option margin, even when the magnitude or uncertainty of the target change matches the original reconstruction (Figure~\ref{fig:hidden_correction_mechanism}a). Similarly, replacing the evidence crop with irrelevant spatial regions degrades the ability to reverse late erroneous preferences (Figure~\ref{fig:evidence_specificity_decision}a). These results show that the student's behavioral change is driven by the exact visual constraints extracted from the intra-teacher contrast rather than an undirected shift in the output distribution.

\Needspace{6\baselineskip}
\paragraph{Post-reflection dynamics of visual recovery.}
We further examine how this evidence-grounded correction unfolds across the generated response after a reflection token. We track both the visual support for subsequent tokens and the decision margin for the correct answer to characterize how the student's continuation and answer preference evolve after self-interruption.

We observe complementary changes in visual support and answer preference after reflection. Relative to Vision-OPD, OPD-Aha's mean correct-option margin advantage is largest in the first 1--8 tokens and remains positive across the later windows (Figure~\ref{fig:evidence_specificity_decision}b). Its visual-support advantage is larger in the later windows than immediately after reflection (Figure~\ref{fig:hidden_correction_mechanism}b). Thus, an early answer-margin advantage accompanies a continuation that increasingly draws on visual evidence. Together, these trends connect self-interruption with renewed visual reliance and improved answer preference along the subsequent generation.

\FloatBarrier
\Needspace{18\baselineskip}

\section{Experimental Configurations and Robustness Evaluations}

\subsection{Experimental Setup}
\label{sec:experimental_setup}

We evaluate \method{} with Qwen3.5 students at the 4B and 9B scales on six fine-grained visual benchmarks: V$^\star$Bench~\citep{wu2024vstar}, ZoomBench~\citep{wei2026zoombench}, HR-Bench-4K and HR-Bench-8K~\citep{wang2025hrbench}, and the English and Chinese splits of MME-RealWorld~\citep{zhang2025mmerealworld}. Our controlled comparison uses the 6,241 training examples released with Vision-OPD~\citep{yuan2026visionopd} and holds the student initialization, rollout and update budgets, and decoding configuration fixed across GRPO~\citep{shao2024deepseekmath}, Vision-OPD, V-Zero, and \method{}. The student generates from the original full image, while methods using privileged supervision receive the same localized evidence crop. \method{} additionally constructs its visual null by replacing this crop with its mean RGB color. General-purpose and agentic multimodal models provide broader capability context. We report per-benchmark accuracy and the unweighted mean across the six benchmarks (Avg$_6$). Full training configurations are provided in Appendix~\ref{sec:implementation_details_appendix}.

\subsection{Training Details}
\label{sec:implementation_details_appendix}
Reconstructing the distillation target requires separating the effect of visual evidence from model-specific differences in capacity and calibration. Within each model scale, the same frozen copy of the initial student serves as the teacher for both the real-evidence and visual-null predictions. The model, shared student prefix, and token positions remain fixed, so the resulting change in token preference is attributable to the localized visual evidence. This correction is distilled along the unchanged student rollout, while inference retains only the student and the original full image. Table~\ref{tab:training_configuration} summarizes the shared optimization configuration.

\begin{table}[H]
\caption{Training configuration for \method{}.}
\label{tab:training_configuration}
\centering
\setlength{\tabcolsep}{10pt}
\begin{tabular}{l l}
\toprule
\textbf{Configuration} & \textbf{Value} \\
\midrule
Student backbone & Qwen3.5-4B / Qwen3.5-9B \\
Teacher & Frozen copy of the initial student \\
Training data & Vision-OPD-6K (6,241 examples) \\
Student visual input & Original full image \\
Teacher real-evidence input & Localized evidence crop \\
Teacher visual-null input & Mean RGB crop with matched dimensions \\
Distillation objective & JSD (4B and 9B) \\
Distribution support & student top-100 tokens and one tail bucket \\
Learning rate & $2\times10^{-6}$ \\
Global batch size & 96 \\
Rollouts per prompt & 8 \\
Maximum prompt / response length & 8,192 / 1,024 tokens \\
Random seed & 42 \\
\bottomrule
\end{tabular}
\end{table}

\FloatBarrier
\Needspace{8\baselineskip}

\subsection{Target Reconstruction Remains Robust Across Supervision Geometries}
\label{sec:supervision_divergence_appendix}

The reconstructed target defines the corrective token distribution, while the supervision divergence determines the optimization geometry used to align the student with that target. Contrasting the symmetric Jensen--Shannon divergence with the directional Forward and Reverse KL alternatives separates the benefit of target reconstruction from a particular loss formulation.

\begin{table}[H]
\caption{Ablation of the supervision divergence at the 4B and 9B scales.}
\label{tab:supervision_divergence_ablation}
\centering
\small
\setlength{\tabcolsep}{4.5pt}
\begin{tabular}{l|cc|cc|cc|c}
\toprule
\textbf{Divergence} & \textbf{V$^\star$} & \textbf{Zoom} & \textbf{HR-4K} & \textbf{HR-8K} & \textbf{MME-EN} & \textbf{MME-CN} & \textbf{Avg$_6$} \\
\midrule
\multicolumn{8}{l}{\textbf{\textit{Qwen3.5-4B}}} \\
Forward KL & \textbf{94.76} & 59.53 & 86.38 & 82.88 & 76.56 & 74.06 & 79.03 \\
Reverse KL & 93.72 & 62.01 & 87.00 & \textbf{84.88} & 76.31 & 73.89 & 79.63 \\
JSD & 93.70 & \textbf{62.50} & \textbf{88.60} & 84.80 & \textbf{77.70} & \textbf{74.90} & \textbf{80.40} \\
\midrule
\multicolumn{8}{l}{\textbf{\textit{Qwen3.5-9B}}} \\
Forward KL & 89.01 & 57.99 & 83.00 & 79.88 & 70.38 & 71.15 & 75.23 \\
Reverse KL & \textbf{94.80} & 60.90 & \textbf{89.50} & 86.50 & 77.90 & \textbf{75.40} & 80.80 \\
JSD & 94.76 & \textbf{63.91} & 88.88 & \textbf{87.50} & \textbf{78.28} & 75.24 & \textbf{81.43} \\
\bottomrule
\end{tabular}
\end{table}

JSD achieves the highest Avg$_6$ at both 4B and 9B, while the best per-benchmark results remain distributed across divergence choices (Table~\ref{tab:supervision_divergence_ablation}). Multiple divergence choices nevertheless retain the gains of target reconstruction, showing that its benefit arises from the reconstructed visual preference rather than a single supervision geometry.

\FloatBarrier
\Needspace{8\baselineskip}

\subsection{Log-Probability Preserves the Relational Structure of Visual Evidence}
\label{sec:reconstruction_geometry_ablation}

Translating the visual preference into a valid training target requires choosing how to represent the evidence-induced prediction shift. Probability-space reconstruction treats this shift as an absolute displacement of probability mass, $q_t^{\mathrm{prob}}(\beta;v)\propto[(1+\beta)p_t^{+}(v)-\beta p_t^{0}(v)]_{+}$. Log-probability reconstruction instead represents the relative change between the real and visual-null predictions, $q_t^{\mathrm{log}}(\beta;v)\propto p_t^{+}(v)\bigl(p_t^{+}(v)/p_t^{0}(v)\bigr)^{\beta}$.

\begin{table}[H]
\caption{Probability-space and log-probability reconstruction yield comparable aggregate gains.}
\label{tab:reconstruction_geometry_ablation}
\centering
\setlength{\tabcolsep}{4.2pt}
\resizebox{\textwidth}{!}{%
\begin{tabular}{l|c|cc|cc|cc|c}
\toprule
\textbf{Target} & \textbf{Representation} & \textbf{V$^\star$} & \textbf{Zoom} & \textbf{HR-4K} & \textbf{HR-8K} & \textbf{MME-EN} & \textbf{MME-CN} & \textbf{Avg$_6$} \\
\midrule
Standard privileged target & --
& 89.0 & 59.5 & 83.4 & 80.5 & 74.6 & 71.6 & 76.4 \\
Probability-space target & Probability
& \textbf{95.8} & 60.8 & 85.3 & \textbf{82.0} & 75.6 & 72.5 & 78.7 \\
\rowcolor{tablehighlight}\textbf{Reconstructed target} & Log-probability
& 93.2 & \textbf{61.5} & \textbf{86.0} & \textbf{82.0} & \textbf{76.5} & \textbf{73.2} & \textbf{78.8} \\
\bottomrule
\end{tabular}
}
\end{table}

Both representations improve over the standard privileged target and produce similar aggregate gains (Table~\ref{tab:reconstruction_geometry_ablation}). The log-probability form more directly preserves the multiplicative relation between the real and visual-null predictions. Absolute probability shifts depend on the initial scale of each token probability, whereas the log-probability formulation retains the privileged teacher distribution as a structural anchor and applies an exponential tilt along the visual preference.

\FloatBarrier
\Needspace{8\baselineskip}

\subsection{Reconstructed Supervision Is Robust to Visual-Null Constructions}
\label{sec:visual_null_ablation}

The visual null removes task-relevant evidence while providing a reference for the same teacher under the shared student prefix. Alternative constructions test whether the corrective signal reflects the active contribution of privileged evidence or an artifact of a particular null input. We consider Gaussian noise, a mismatched natural image, a black image, and the mean-color transformation used by \method{}.

\begin{table}[H]
\caption{Ablation of the visual null.}
\label{tab:visual_null_ablation}
\centering
\setlength{\tabcolsep}{4.7pt}
\resizebox{\textwidth}{!}{%
\begin{tabular}{l|cc|cc|cc|c}
\toprule
\textbf{Visual null $I^0$} & \textbf{V$^\star$} & \textbf{Zoom} & \textbf{HR-4K} & \textbf{HR-8K} & \textbf{MME-EN} & \textbf{MME-CN} & \textbf{Avg$_6$} \\
\midrule
Gaussian noise
& \textbf{95.3} & 61.0 & 86.5 & 83.8 & 77.3 & 74.9 & 79.8 \\
Mismatched image
& 94.8 & \textbf{63.0} & 85.9 & 82.9 & 77.0 & 74.1 & 79.6 \\
Black image
& \textbf{95.3} & 62.5 & 87.6 & 83.6 & \textbf{77.8} & \textbf{75.1} & 80.3 \\
\rowcolor{tablehighlight}\textbf{Mean RGB color (Ours)}
& 93.7 & 62.5 & \textbf{88.6} & \textbf{84.8} & 77.7 & 74.9 & \textbf{80.4} \\
\bottomrule
\end{tabular}
}
\end{table}

The extracted visual preference produces similar aggregate improvements across all four constructions (Table~\ref{tab:visual_null_ablation}). This stability shows that target reconstruction depends on removing the privileged visual content rather than on the specific appearance of the null input. The mean-color transformation therefore provides a simple and effective default without being a critical design choice.

\FloatBarrier
\Needspace{8\baselineskip}

\subsection{Training Dynamics}
\label{sec:training_dynamics_appendix}

At each update, we track the token-level JSD between the student and the reconstructed target, the actor gradient norm, and answer accuracy on generated rollouts. These trajectories connect the optimization process to changes in generation and final decisions.

\begin{figure}[!htbp]
    \centering
    \includegraphics[width=0.98\textwidth]{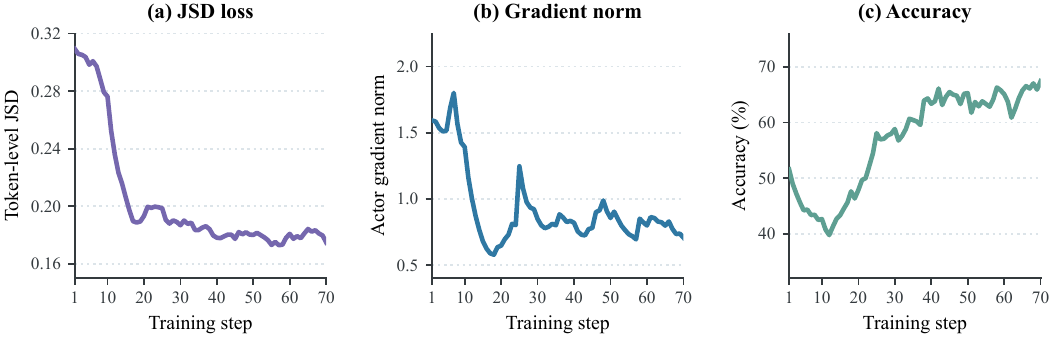}
    \caption{Training dynamics of the 4B student with $\beta=4$. \textbf{(a)} Token-level JSD falls sharply early in training and then stabilizes. \textbf{(b)} The actor gradient norm follows the same transition before settling into a lower range. \textbf{(c)} Answer accuracy on the generated rollouts initially decreases and then recovers.}
    \label{fig:training_dynamics_beta4}
\end{figure}

The JSD and gradient activity decrease before accuracy begins to recover. This early interval overlaps with the transient response-length increase in Figure~\ref{fig:revision_mechanism_sequence}c. The reconstructed target therefore changes generation before improved decisions become visible in rollout accuracy.

\FloatBarrier
\Needspace{8\baselineskip}

\subsection{Qualitative Case Studies}
\label{sec:qualitative_case_studies_appendix}

\begin{figure}[p]
    \centering
    \includegraphics[page=1,width=0.96\textwidth]{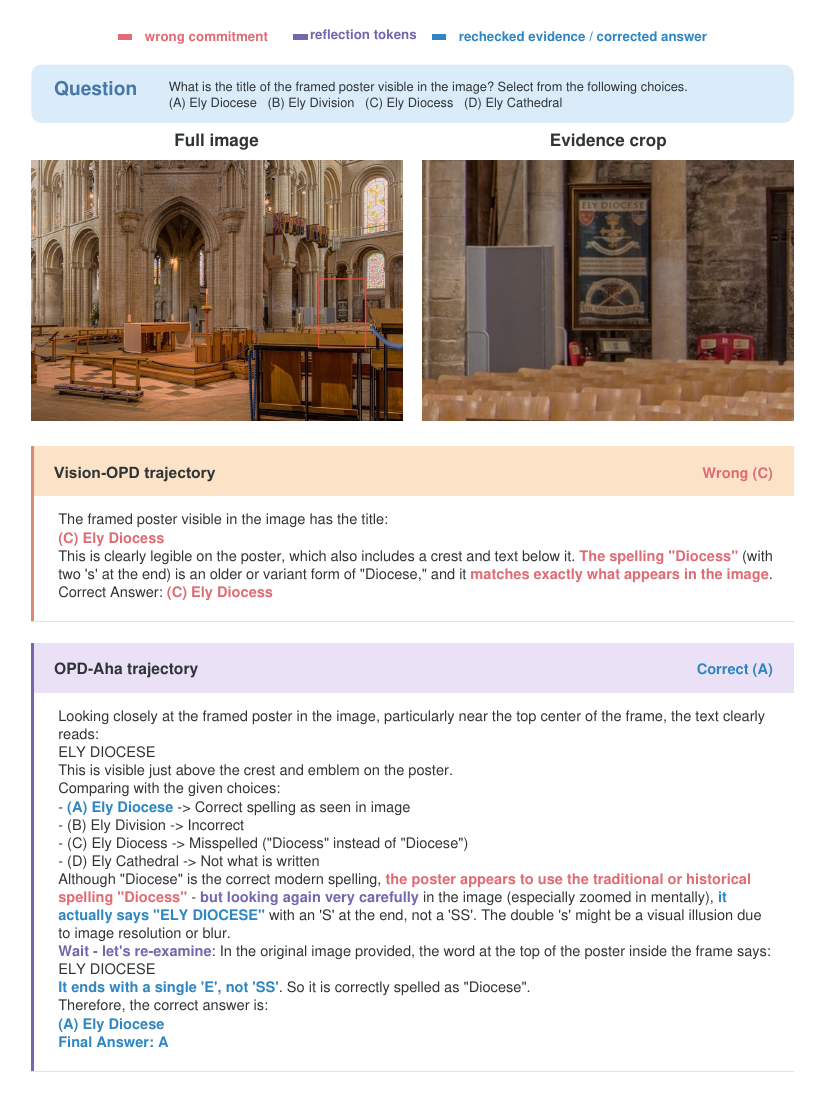}
    \caption{\textbf{Fine-grained text recognition.} Vision-OPD commits to option C by reading the poster title as ``Ely Diocess'' and rationalizing the extra ``s.'' \method{} initially reads ``ELY DIOCESE'' correctly, then questions the spelling and considers ``Diocess.'' After reflection, it re-examines the poster text, returns to its original reading, and selects option A. The trajectory shows a renewed check of the diagnostic letter sequence after an intervening mistaken interpretation.}
    \label{fig:case_study_text}
\end{figure}

\begin{figure}[p]
    \centering
    \includegraphics[page=2,width=0.88\textwidth]{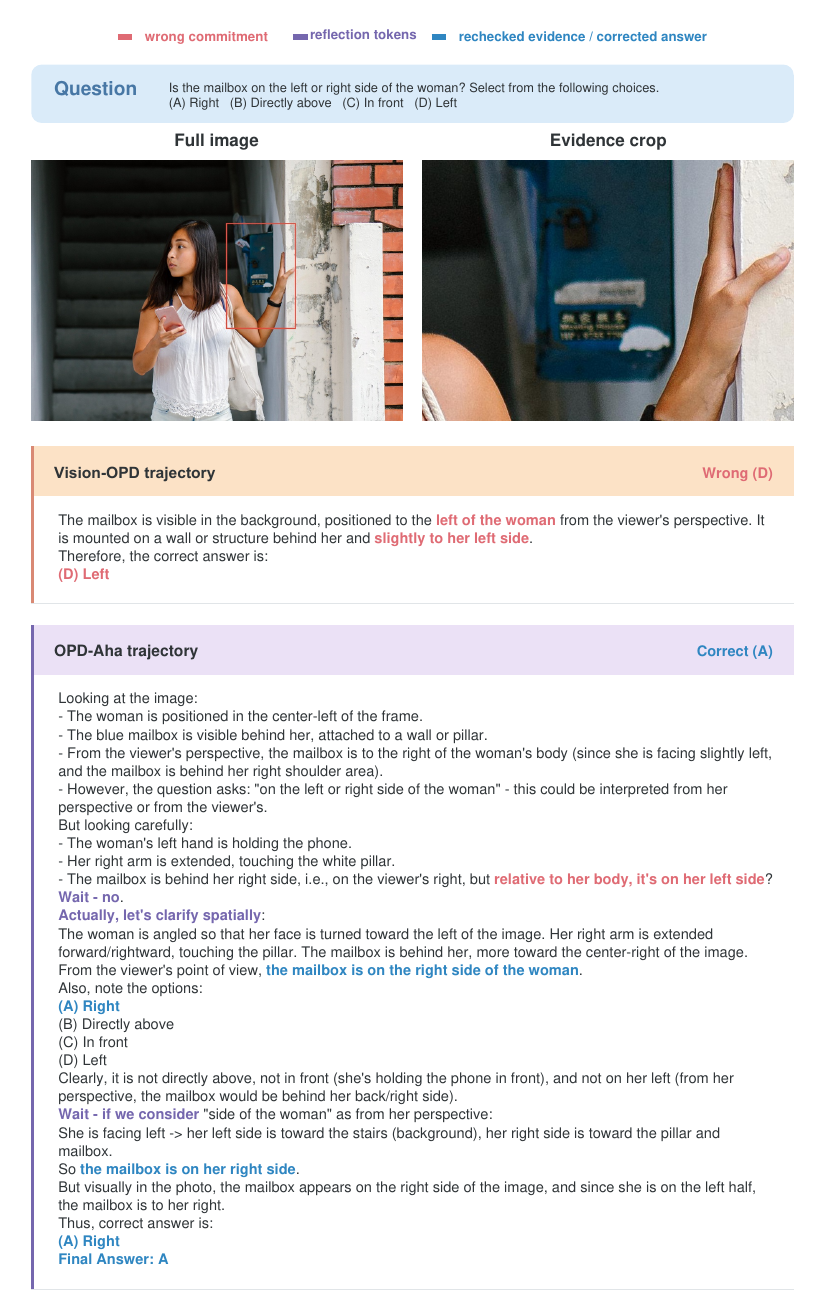}
    \caption{\textbf{Spatial viewpoint resolution.} Vision-OPD anchors on the mailbox's horizontal image position and selects option D, placing it on the woman's left. \method{} questions the reference frame, relates the mailbox to the woman's body and outstretched right arm, and selects option A. During reflection, it rechecks both the woman's pose and the mailbox's image position to resolve the left--right ambiguity.}
    \label{fig:case_study_spatial}
\end{figure}

\begin{figure}[p]
    \centering
    \includegraphics[page=3,width=0.94\textwidth]{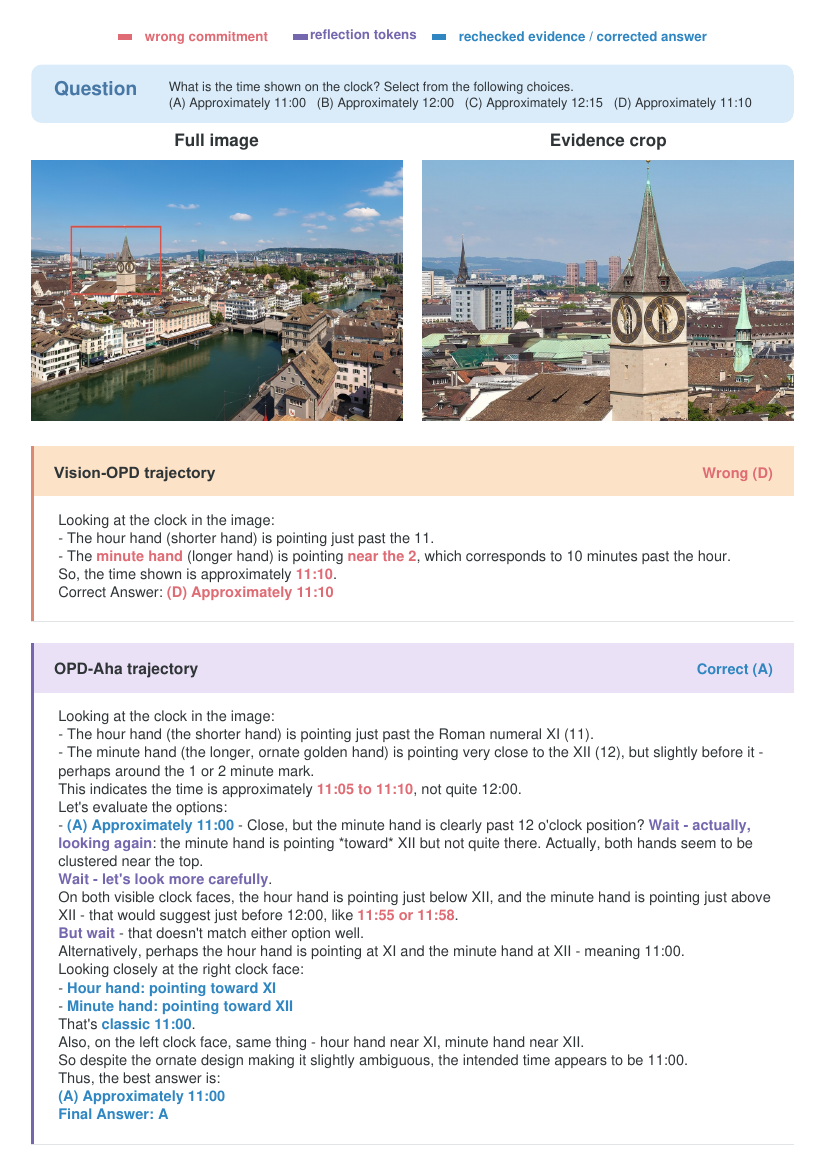}
    \caption{\textbf{Distant clock reading.} Vision-OPD misreads the minute hand as pointing near II and selects approximately 11:10, option D. \method{} revisits both visible clock faces, distinguishes the short hour hand near XI from the long minute hand near XII, and selects approximately 11:00, option A. Cross-checking the repeated visual evidence prevents one ambiguous hand estimate from determining the final answer.}
    \label{fig:case_study_clock}
\end{figure}

We select three paired cases from HR-Bench-4K and HR-Bench-8K to examine how the reconstructed target changes generation beyond the final accuracy score. In each case, Vision-OPD produces an incorrect answer while \method{} answers the same question correctly. Figures~\ref{fig:case_study_text}, \ref{fig:case_study_spatial}, and \ref{fig:case_study_clock} show the full image, the localized evidence crop, and the complete generated trajectories from both methods. Red marks the interpretation that supports the initial wrong commitment, purple marks the reflection that interrupts this continuation, and blue marks the rechecked visual evidence and corrected answer. The cases cover fine-grained text recognition, spatial viewpoint resolution, and distant clock reading.

Across the three cases, \method{} revisits uncertain interpretations before committing to the final answer. The student recognizes that its current interpretation may be unreliable, then returns to the diagnostic region and rechecks the decision against visual evidence. The reflection token marks the pivot between these stages: it interrupts the uncertain continuation and precedes a renewed check of the visual evidence. Vision-OPD instead remains consistent with its first reading and turns an early perceptual mistake into a confident wrong answer. The trajectories illustrate how reconstructed supervision can lead from self-interruption to renewed use of visual evidence.

\end{document}